\documentclass[10pt,twocolumn,letterpaper]{article}

\usepackage{cvpr}
\usepackage{times}
\usepackage{epsfig}
\usepackage{graphicx}
\usepackage{amsmath}
\usepackage{amssymb}
\usepackage{booktabs}
\usepackage{subfig}

\usepackage{multirow} 
\usepackage{array}
\usepackage{enumitem}

\newcommand{\x}{$\times$}

\usepackage[pagebackref=true,breaklinks=true,letterpaper=true,colorlinks,bookmarks=false]{hyperref}

\cvprfinalcopy 

\def\cvprPaperID{2998} 

\begin{document}

\title{Logo Synthesis and Manipulation with Clustered Generative Adversarial Networks}

\author{Alexander Sage\\
D-ITET, ETH Zurich\\
Switzerland\\
{\tt\small ~~~~~~~sagea@ee.ethz.ch~~~~~~~}
\and
Eirikur Agustsson\\
D-ITET, ETH Zurich\\
Switzerland\\
{\tt\small aeirikur@vision.ee.ethz.ch}
\and
Radu Timofte\\
D-ITET, ETH Zurich\\
Merantix GmbH\\
{\tt\small radu.timofte@vision.ee.ethz.ch}
\and
Luc Van Gool\\
D-ITET, ETH Zurich\\
ESAT, KU Leuven\\
{\tt\small vangool@vision.ee.ethz.ch}
}
\maketitle

\begin{abstract}
Designing a logo for a new brand is a lengthy and tedious back-and-forth process between a designer and a client. In this paper we explore to what extent machine learning can solve the creative task of the designer.
For this, we build a dataset -- LLD -- of 600k+ logos crawled from the world wide web. Training Generative Adversarial Networks (GANs) for logo synthesis on such multi-modal data is not straightforward and results in mode collapse for some state-of-the-art methods. We propose the use of synthetic labels obtained through clustering to disentangle and stabilize GAN training. We are able to generate a high diversity of plausible logos and we demonstrate latent space exploration techniques to ease the logo design task in an interactive manner. Moreover, we validate the proposed clustered GAN training on CIFAR 10, achieving state-of-the-art Inception scores when using synthetic labels obtained via clustering the features of an ImageNet classifier. GANs can cope with multi-modal data by means of synthetic labels achieved through clustering, and our results show the creative potential of such techniques for logo synthesis and manipulation. Our dataset and models will be made publicly available at \url{https://data.vision.ee.ethz.ch/cvl/lld/}.
\end{abstract}

\section{Introduction and related work}
\label{sec:intro}

\paragraph{Logo design}Designing a logo for a new brand usually is a lengthy and tedious process, both for the client and the designer. A lot of ultimately unused drafts are produced, from which the client selects his favorites, followed by multiple cycles refining the logo to match the clients needs and wishes. Especially for those clients without a specific idea of the end product, this results in a procedure that is not only time, but also cost intensive.

\begin{figure}[tbp]
	\centering
	\begin{tabular}{cccc}
		\includegraphics[width=0.1\textwidth]{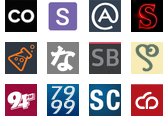} &
		\includegraphics[width=0.1\textwidth]{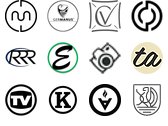} &
		\includegraphics[width=0.1\textwidth]{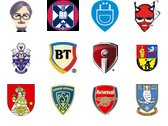} &
		\includegraphics[width=0.1\textwidth]{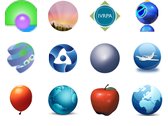} \\
		\includegraphics[width=0.1\textwidth]{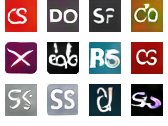} &
		\includegraphics[width=0.1\textwidth]{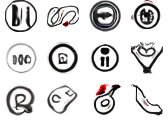} &
		\includegraphics[width=0.1\textwidth]{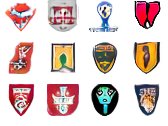} &
		\includegraphics[width=0.1\textwidth]{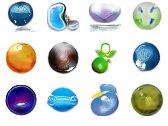}
	\end{tabular}
	\vspace{-0.2cm}
	\caption{Original and generated images from four selected clusters from our LLD-icon-sharp dataset. The top three rows consist of original logos, followed by logos generated using our iWGAN-LC trained on 128 RC clusters.}
	\label{fig:teaser}
	\vspace{-0.2cm}
\end{figure}

\begin{figure*}[tb]
	\centering
	\includegraphics[width=0.8\textwidth]{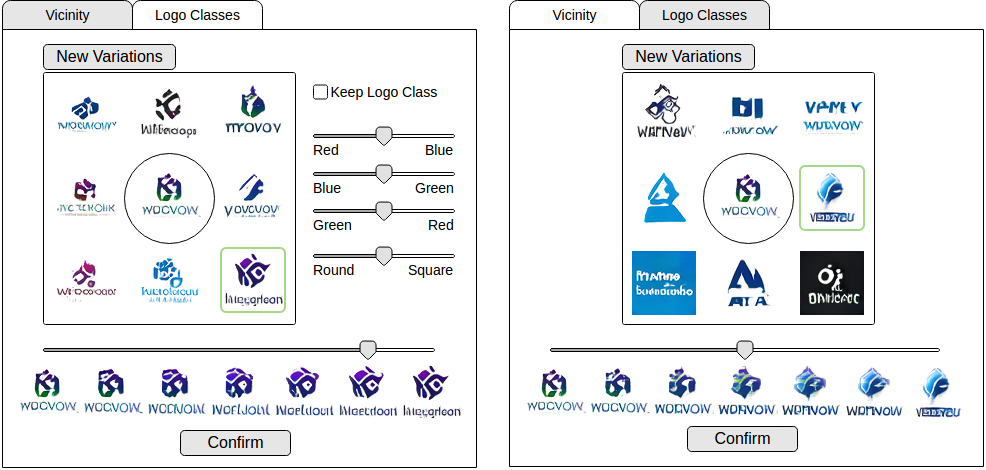}
	\caption{Logo generator interface. The user is able to choose either vicinity sampling or class transfer to modify the image in a chosen semantic direction. For both methods, 8 random variations are arranged around the current logo. Upon selecting the appropriate sample, the current logo can be modified by a variable amount using the slider at the bottom of the window. After confirming the selected modification, the process starts over again from the newly modified logo, until the desired appearance is reached. In addition to vicinity sampling within or across clusters, some pre-defined semantic modifications can be made using the sliders on the right hand side of the first view. The images used here are generated with iWGAN-LC trained  at 64\x64 pixels on LLD-logo clustered to 64 different classes as explained in Section~\ref{sec:clustered_GAN}.}
	\label{fig:interface}
\end{figure*}

The goal of this work is to provide a framework towards a system with the ability to generate (virtually) infinitely many variations of logos (some examples are shown in Figure~\ref{fig:teaser}) to facilitate and expedite such a process. To this end, the prospective client should be able to modify a prototype logo according to specific parameters like shape and color, or shift it a certain amount towards the characteristics of another prototype.  An example interface for such a system is presented in Figure~\ref{fig:interface}. It could help both designer and client to get an idea of a potential logo, which the designer could then build upon, even if the system itself was not (yet) able to output production-quality designs.

\paragraph{Logo image data}Existing research literature focused mostly on retrieval, detection, and recognition of a reduced number of logos~\cite{LOGO-Net-15,joly2009logo,romberg2011scalable,sahbi2013context,Su_2017_ICCV,zhu2007automatic} and, consequently, a number of datasets were introduced. The most representative large public logo datasets are shown in Table~\ref{tab:logo_datasets}. Due to the low diversity of the contained logos, these datasets are not suitable for learning and validating automatic logo generators. At the same time a number of web pages allow (paid) access to a large number of icons, such as iconsdb.com (4135+ icons), icons8.com (59900+), iconfinder.com (7473+), iconarchive.com (450k+) and thenounproject.com (1m+). However, the diversity of these icons is limited by the number of sources, namely designers/artists, themes (categories) and design patterns (many are black and white icons). Therefore, we crawl a highly diverse dataset -- the Large Logo Dataset (LLD) -- of real logos `in the wild' from the Internet. As shown in Table~\ref{tab:logo_datasets} our LLD proposes thousands of times more distinct logos than the largest public logo dataset to date, WebLogo-2M~\cite{Su_2017_ICCV}.

In contrast to popularly used natural image datasets such as ImageNet~\cite{imagenet}, CIFAR-10~\cite{CIFAR10} and LSUN~\cite{yu2015lsun}, face datasets like CelebA~\cite{liu2015faceattributes} and the relatively easily modeled handwritten digits of MNIST~\cite{lecun1998gradient}, logos are: (1) \textit{Artificial}, yet strongly multimodal and thus challenging for generative models; (2) \textit{Applied}, as there is an obvious real-world demand for synthetically generated, unique logos since they are expensive to produce; (3) \textit{Hard to label}, as there are very few categorical properties which manifest themselves in a logo's visual appearance.
While the logos are easily obtainable in large quantities, they are specifically designed to be unique, which ensures the diversity of a large logo dataset. We argue that all these characteristics make logos a very attractive domain for machine learning research in general, and generative modeling in particular.

\paragraph{Generative models}Recent advances in generative modeling have provided viable frameworks for making such a system possible. The current state-of-the-art is made up mainly of two types of generative models, namely Variational Autoencoders (VAEs)~\cite{jimenez2014stochastic,kingma2014semi,kingma2013auto} and Generative Adversarial Networks (GANs)~\cite{arjovsky2017towards,goodfellow2016nips,goodfellow2014generative}. Both of these models generate their images from a high-dimensional latent space that can act as a sort of ``design space'' in which a user is able to modify the output in a structured way. VAEs have the advantage of directly providing embeddings of any given image in the latent space, allowing targeted modifications to its reconstruction, but tend to suffer from blurry output owed to the nature of the pixel-wise $L_2$ loss used during training. GANs on the other hand, which consist of a separate generator and discriminator network trained simultaneously on opposing objectives in a competitive manner, are known to provide realistic looking, crisp images but are notoriously unstable to train. To address this difficulty, a number of improvements in the architecture and training methods of GANs have been suggested~\cite{salimans2016improved}, such as using deep convolutional layers~\cite{radford2015unsupervised} or modified loss functions e.g.\@ based on least-squares~\cite{mao2016least} or the Wasserstein distance between probability distributions~\cite{arjovsky2017WGAN,berthelot2017began,gulrajani2017improvedWGAN}.

\paragraph{Conditional models}The first extension of GANs with class-conditional information~\cite{mirza2014conditional} followed shortly after its inception, generating MNIST digits conditioned on class labels provided to both generator and discriminator during training. It has since been shown for supervised datasets, that class-conditional variants of generative networks very often produce superior results compared to their unconditional counterparts \cite{gulrajani2017improvedWGAN,huang2016stacked,ACGAN}. By adding an encoder to map a real image into the latent space, it was proven to be feasible to generate a modified version of the original image by changing class attributes on faces~\cite{brock2016neural, perarnau2016invertible} and other natural images~\cite{van2016conditional}. Other notable applications include the generation of images from a high-level description such as various visual attributes~\cite{yan2016attribute} or text descriptions~\cite{reed16}.

\paragraph{Our contributions}In this work we train GANs on our own highly multi-modal logo data as a first step towards user-manipulated artificial logo synthesis. Our main contributions are:
\begin{itemize}[]
	\item LLD - a novel dataset of 600k+ logo images.
	\item Methods to successfully train GAN models on multi-modal data. Our proposed clustered GAN training achieves state-of-the-art Inception scores on the CIFAR10 dataset.
	\item An exploration of GAN latent space for logo synthesis.
\end{itemize}

The remainder of this paper is structured as follows. We introduce a novel Large Logo Dataset (LLD) in Section~\ref{sec:lld}. We describe the proposed clustered GAN training, the clustering methods, as well as the GAN architectures used and perform quantitative experiments in Section~\ref{sec:clustered_GAN}. Then we demonstrate logo synthesis by latent space exploration operations in Section~\ref{sec:logo_synthesis}. Finally, we draw the conclusions in Section~\ref{sec:conclusions}.

\begin{table}
	\centering
	\begin{tabular}{l|rr}
		Dataset&Logos&Images\\
		\hline\hline
		FlickLogos-27~\cite{kalantidis2011scalable}&27&1080\\
		FlickLogos-32~\cite{romberg2011scalable}&32&8240\\
		BelgaLogos~\cite{joly2009logo}&37 & 10000\\
		LOGO-Net~\cite{LOGO-Net-15}&160&73414\\
		WebLogo-2M~\cite{Su_2017_ICCV}& 194&1867177\\
		\hline
		\textbf{LLD-icon (ours)}&486377 & 486377\\
		\textbf{LLD-logo (ours)}&122920 & 122920\\
		\textbf{LLD (ours)}     &486377+ & 609297\\
	\end{tabular}
	\caption{Logo datasets. Our LLD provides orders of magnitude more logos than the existing public datasets.}
	\label{tab:logo_datasets}
	\vspace{-0.3cm}
\end{table}

\section{LLD: Large Logo Dataset}
\label{sec:lld}
In the following we introduce a novel dataset based on website logos, called the Large Logo Dataset (LLD). It is the largest logo dataset to date (see Table~\ref{tab:logo_datasets}). The LLD dataset consists of two parts, a low resolution (32\x32 pixel) favicon subset (LLD-icon) and the higher-resolution (400\x400 pixel) twitter subset (LLD-logo). In the following we will briefly describe the acquisition, properties and possible use-cases for each. Both versions will be made available at \url{https://data.vision.ee.ethz.ch/cvl/lld/}.

\subsection{LLD-icon: Favicons}
\begin{figure}[htbp]
	\centering
	\includegraphics[width=\linewidth]{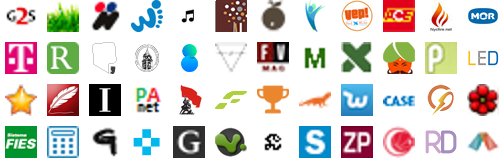}
	\vspace{-0.4cm}
	\caption{Excerpt from LLD-icon.}
	\label{fig:favicon-data}
\end{figure}
For generative models like GANs, the difficulty of keeping the network stable during training increases with image resolution. Thus, when starting to work with a new type of data, it makes sense to start off with a variant which is inherently low-resolution. Luckily, in the domain of logo images there is a category of such inherently low-resolution, low-complexity images: Favicons, the small icons representing a website e.g.\@ in browser tabs or favorite lists. We decided to crawl the web for such favicons using the largest resource of high quality website URLs we could find: Alexa's top 1-million website list\footnote{now officially retired, formerly available at \url{https://www.alexa.com}}. To this end we use the Python package Scrapy\footnote{\url{https://scrapy.org/}} in conjunction with our own download script which directly converts all icons found to a standardized $32\times32$ pixel resolution and RGB color space, discarding all non-square images.

After acquiring the raw data from the web, we remove all exact duplicates (of which there are a surprisingly high number of almost 20~\%). Visual inspection of the raw data reveals a non-negligible number of images that do not comply to our initial dataset criteria and often are not even remotely logo-like, such as faces and other natural images. In an attempt to get rid of this unwanted data, we (i) sort all images by PNG-compressed file size -- an image complexity indicator; (ii) manually inspect and partition the resulting sorted list into three sections: clean and mostly clean data which are kept, and mostly unwanted data which is discarded; (iii) discard the mostly clean images containing the least amount of white pixels.

The result of this process, a small sample of which is given in Figure~\ref{fig:favicon-data}, is a clean set of 486,377 images of uniform 32\x32 pixel size, making it very easy to use. The disadvantage of this standardized size is that 54~\% of images appear blurry because they where scaled up from a lower resolution. For this reason we will also be providing (the indices for) a subset of the data containing only sharp images, which we will refer to as icons-sharp.

\subsection{LLD-logo: Twitter}
\label{sec:lld-logo}
\begin{figure}[htbp]
	\centering
	\includegraphics[width=\linewidth]{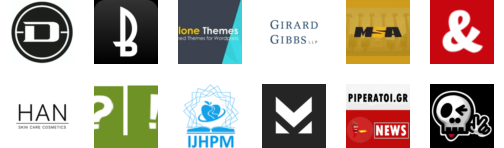}
	\vspace{-0.4cm}
	\caption{Excerpt from LLD-logo, scaled down to 64\x64 pixels.}
	\label{fig:twitter-data}
\end{figure}
For training generative networks at an increased resolution, additional high-resolution data is needed, which favicons cannot provide. One possible option would be to crawl the respective websites directly to look for the website or company logo. However, (a) it might not always be straight-forward to find the logo and distinguish it from other images on the website and (b) the aspect ratio and resolution of logos obtained in this way will be very varied, which would necessitate extensive cropping and resizing, potentially degrading the quality of a large portion of logos.

By crawling twitter instead of websites, we are able to acquire standardized square 400\x400 pixel profile images which can easily be downloaded through the twitter API without the need for web scraping. We use the Python wrapper tweepy to search for the (sub-) domain names contained in the alexa list and match the original URL with the website provided in the twitter profile to make sure that we have found the right twitter user. The images are then run through a face detector to reject any personal twitter accounts and the remaining images are saved together with the twitter meta data such as user name, number of followers and description. For this part of the dataset, all original resolutions are kept as-is, where 80\% are at 400\x400 pixels and the rest at some lower resolution (details given in supplementary material).

The acquired images are analyzed and sorted with a combination of automatic and manual processing in order to get rid of unwanted and possibly sensitive images, resulting in 122,920 usable high-resolution logos of consistent quality with rich meta data from the respective twitter accounts. These logo images form the LLD-logo dataset, a small sample of which is presented in Figure~\ref{fig:twitter-data}.

\section{Clustered GAN Training}
\label{sec:clustered_GAN}
We propose a method for stabilizing GAN training and achieving superior quality samples on unlabeled datasets by means of clustering (a) in the latent space of an autoencoder trained on the same data or (b) in the CNN feature space of a ResNet classifier trained on ImageNet. With both methods we are able to produce semantically meaningful clusters that improve GAN training.

In this Section we review the GAN architectures used in our study, describe the clustering methods based on Autoencoder latent space and ResNet features and discuss the quantitative experimental results.

\subsection{GAN architectures}
Our generative models are based on Deep Convolutional Generative Adversarial Networks (DCGAN) of Radford~\etal~\cite{radford2015unsupervised} and improved Wasserstein GAN with gradient penalty (iWGAN) as proposed by Gulrajani~\etal~\cite{gulrajani2017improvedWGAN}.
\paragraph{DCGAN} For our DCGAN experiments, we use Taehoon Kim's TensorFlow implementation~\footnote{\url{https://github.com/carpedm20/DCGAN-tensorflow}}.
We train DCGAN exclusively on the low-resolution LLD-icon subset, for which it proved to be inherently unstable without using our clustering approach.
We use the input blurring explained in the next section in all our DCGAN experiments. For details on hyper-parameters used, we refer the interested reader to the supplementary material.

\paragraph{iWGAN} All our iWGAN experiments are based on the official TensorFlow repository by Gulrajani~\etal~\cite{gulrajani2017improvedWGAN}\footnote{\url{https://github.com/igul222/improved_wgan_training}}. We kept the default settings as provided by the authors. We exclusively use the 32- and 64-pixel ResNet architectures provided in the repository with the only major modifications being our conditioning method as described below. We also use linear learning rate decay (from the initial value to zero over the full training iterations) for all our experiments.

\subsection{Clustering}
\label{ssc:clustering}

As mentioned in the introduction (Section~\ref{sec:intro}), training a conditional GAN with labels is beneficial in terms of improved output quality over an unsupervised setting. In particular, we found DCGAN to be unstable with our icon dataset (LLD-icon) for resolutions higher than 10\x10, and where able to stabilize it by introducing synthetic labels as described in this section. In addition to stabilizing GAN training, we are able to achieve state-of-the-art Inception scores (as proposed by Salimans~\etal~\cite{salimans2016improved}) on CIFAR-10 using iWGAN with our synthetic labels produced by RC clustering as described below, and thus demonstrate quantitative evidence of a quality improvement using this approach in Section~\ref{ssc:quantitative_evaluation}.
Furthermore, the cluster labels subsequently allow us to have some additional control over the generated logos by generating samples from individual clusters or transforming an particular logo to inherit the specific attributes of another cluster as demonstrated in Section~\ref{sec:logo_synthesis}.

\begin{figure}[tbp]
\centering
	\def\svgwidth{0.8\linewidth}
	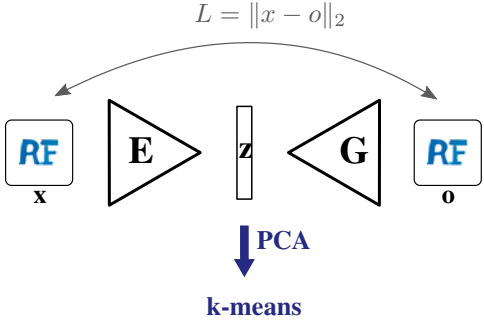
	\caption{Autoencoder used for AE clustering. The generator G is equivalent to the one used in the GAN, while the encoder E consists of the GAN discriminator D with a higher number of outputs to match the dimensionality of the latent space z. It is trained using a simple $L_2$ loss function.}
	\label{fig:autoencoder}
\end{figure}

\paragraph{AE: AutoEncoder Clustering}
Our first proposed method for producing synthetic data labels is by means of clustering in the latent space $z$ of an Autoencoder. We construct an Autoencoder, consisting of a modified version of the GAN discriminator with $\text{dim}(z)$ outputs instead of one, acting as an encoder to latent space, and the unmodified GAN Generator acting as a decoder for the reconstruction of the image from the latent representation, as illustrated in Figure~\ref{fig:autoencoder}. This Autoencoder is trained using a simple $L_2$ loss between original and reconstructed image. All images are then encoded to latent vectors, followed by a PCA dimensionality reduction and finally clustered using (mini-batch) k-means.
For our logo data, this produces clusters that are both semantically meaningful, as they are based on high-level AE features, and recognizable by the GAN because they where created using the same general network topology.

\paragraph{RC: ResNet Classifier Clustering}
For our second clustering method we leverage the learned features of an ImageNet classifier, namely ResNet-50 by He~\etal~\cite{he2016resnet}. We feed our images to the classifier and extract the output of the final pooling layer from the network to get a 2048-dimensional feature vector. After a PCA dimensionality reduction we can cluster our data in this feature space with (minibatch) k-means. The obtained clusters are considerably superior to those produced with our AE clustering method on CIFAR-10, where one could argue that we are benefiting from the similarity in categories between ImageNet and CIFAR-10, and are thus indirectly using labeled data. However, the clustering is very meaningful also on our logo dataset, which has a very different content and does not consist of natural images like ImageNet, proving the generality of this approach.

\subsection{Conditional GAN Training Methods}
\label{sec:conditional_GAN}
In this section we describe the conditional GAN models used to leverage our synthetic data labels and the input blurring applied to DCGAN.

\paragraph{LC: Layer Conditional GAN}

\begin{figure}[b]
	\centering
    \includegraphics[width=0.5\linewidth]{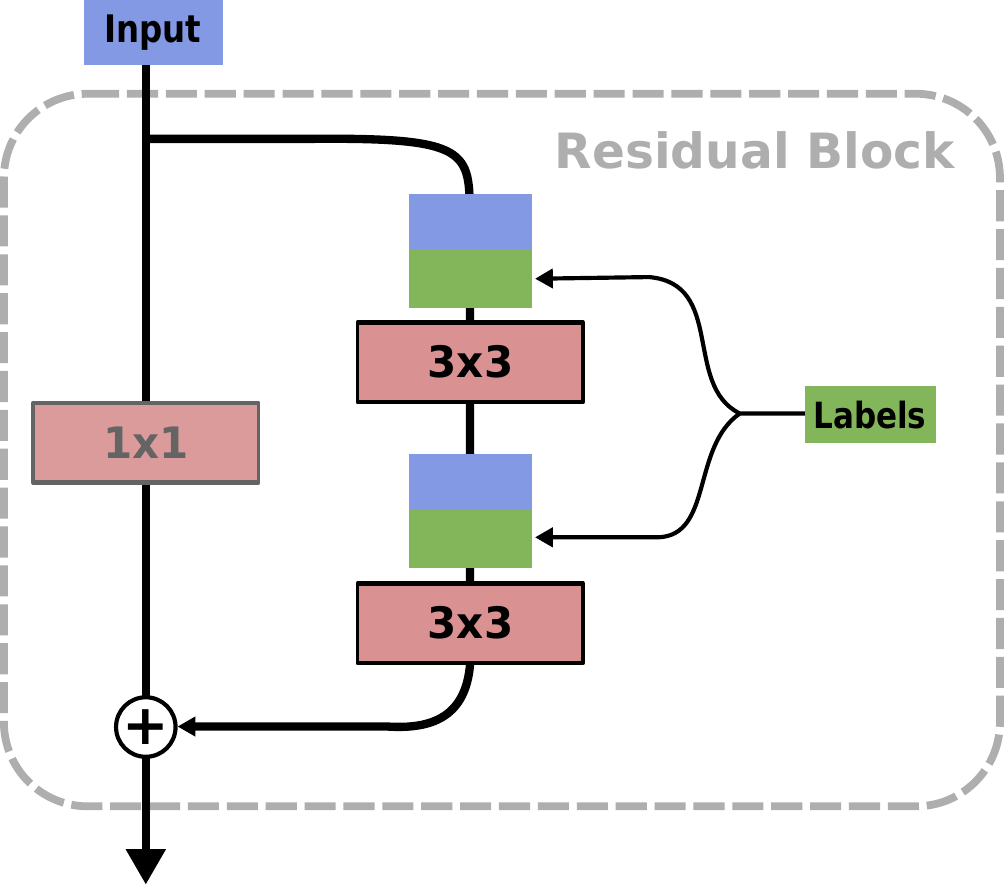}
    \vspace{-0.1cm}
	\caption{Layer Conditional Residual block as used in our iWGAN-LC. The label information is appended to the convolutional layer input in the same way as described in Figure \ref{fig:cond_DCGAN}. The skip connections remain unconditional.}
	\label{fig:cond_resnet}
    \vspace{-0.3cm}
\end{figure}
\begin{figure*}[htbp]
	\centering
	\includegraphics[width=0.8\linewidth]{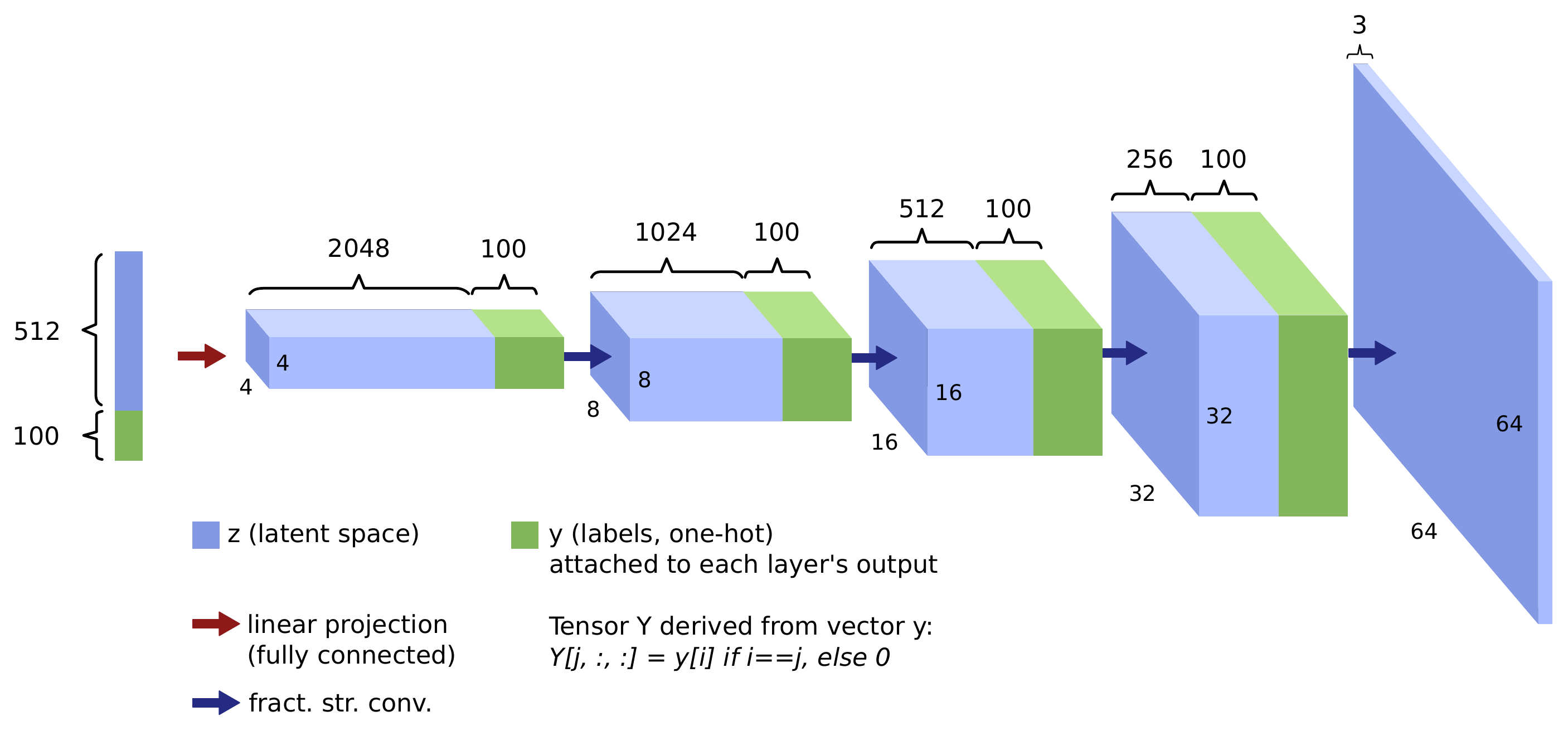}
	\hspace{10px}
	\caption{Generator network as used for our layer conditional DCGAN (DCGAN-LC). 100 labels y are appended as a one-hot vector to the latent vector. It is also projected onto a set of feature maps consisting of all zeros except for the map corresponding to the class number, where all elements have value one. These additional feature maps are then appended to the input of each convolutional layer.}
	\label{fig:cond_DCGAN}
\end{figure*}

In our layer-conditional models, the cluster label for each training sample is fed to all convolutional and linear layers of both generator and discriminator. For linear layers it is simply appended to the input as a one-hot vector. For convolutional layers the labels are projected onto ``one-hot feature maps'' with as many channels as there are clusters, where the one corresponding to the cluster number is filled with ones, while the rest are zero. These additional feature maps are appended to the input of every convolutional layer, such that every layer can directly access the label information. This is illustrated in Figure~\ref{fig:cond_DCGAN} for DCGAN and Figure~\ref{fig:cond_resnet} for ResNet as used in our iWGAN model. Even though the labels are provided to every layer, there is no explicit mechanism forcing the network to use this information. In case the labels are random or meaningless, they can simply be ignored by the network. However, as soon as the discriminator starts adjusting its criteria for each cluster, it forces the generator to produce images that comply with the different requirements for each class. Our experiments confirm that visually meaningful clusters are always picked up by the model, while the network simply falls back to the unconditional state for random labels. 
This type of class conditioning has some useful properties such as the ability to interpolate between different classes and is less prone to failure in producing class-conditional samples compared to the AC conditioning described below. However, it does come with the drawback of adding a significant number of parameters, especially to low-resolution networks, when there are a large number of classes. This effect diminishes with larger networks containing more feature maps, as the number of added parameters remains constant.

\paragraph{AC: Auxiliary Classifier GAN}
With iWGAN we also use the Auxiliary Classifier proposed by Odena~\etal~\cite{ACGAN} as implemented by Glurajani~\etal~\cite{gulrajani2017improvedWGAN}. While this method does not allow us to interpolate between clusters and is thus slightly more limited from an application perspective, it does avoid adding parameters to the convolutional layers, which in general results in a network with fewer parameters. iWGAN-AC was our method of choice for CIFAR-10, as it delivers the highest Inception scores.

\begin{figure}[thbp]
	\centering
	\includegraphics[width=\linewidth]{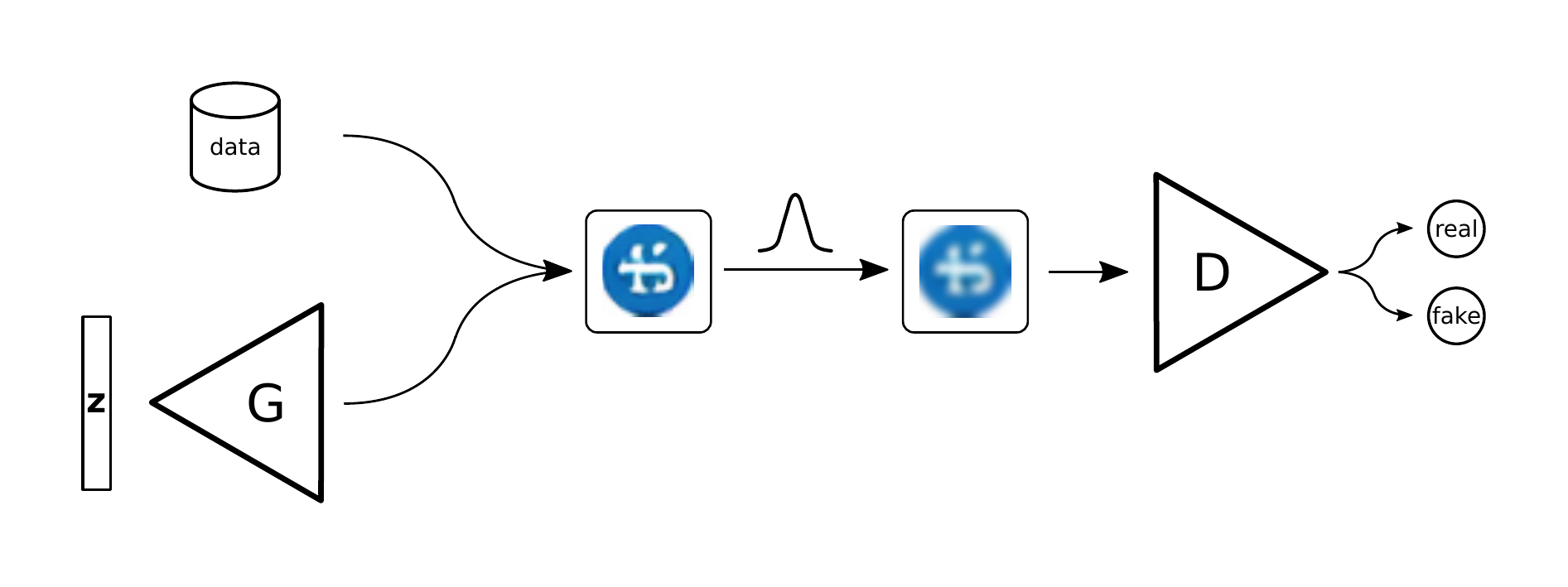}
	\vspace{-0.7cm}
	\caption{Generative Adversarial Net with blurred Discriminator input. Both original and generated images are blurred using a Gaussian filter of fixed strength.}
	\vspace{-0.1cm}
	\label{fig:blurGAN}
\end{figure}

\paragraph{Gaussian Blur}
During our experiments we noticed how blurring the input image helps the network remain stable during training, which in the end lead us to apply a Gaussian blur on all images presented to the discriminator (training data as well as samples from the Generator), like it has been previously implemented by Susmelj~\etal~\cite{ABC-GAN}. The method is schematically illustrated in Figure~\ref{fig:blurGAN}. Upscaling the images to 64$\times$64 pixel resolution before convolving them with the Gaussian kernel enables us to train with blurred images while preserving almost all of the image's sharpness when scaled back down to the original resolution of 32$\times$32 pixels. When generating image samples from the trained Generator without applying the blur filter, there is some noticeable noise in the images, which becomes imperceptible after resizing to the original data resolution while producing almost perfectly sharp output images. Based on our experimental experience we believe this to produce higher quality samples and help stability, it is however not strictly necessary to achieve stability with DCGAN when using clustered training.

\subsection{Quantitative evaluation and state-of-the-art}
\label{ssc:quantitative_evaluation}

In order to quantitatively assess the performance of our solutions on the commonly used CIFAR-10 dataset we report Inception scores~\cite{salimans2016improved} and diversity scores based on MS-SSIM~\cite{MS-SSIM} as suggested in~\cite{ACGAN} over a set of 50000 randomly generated images. In Table~\ref{tab:inception_CIFAR-10} we summarize results for different configurations in supervised (using CIFAR class labels) and unsupervised settings in LC and AC conditional modes, including reported scores from the literature.

\begin{table}[th]
	\centering
	\resizebox{\linewidth}{!}
	{
		\begin{tabular}{clr|c|c}
			\toprule
			&\multirow{2}{*}{\textbf{Method}}&\multirow{2}{*}{\textbf{Clusters}} & \textbf{Inception}& \textbf{Diversity}\\
			&   &   & \textbf{score} & \textbf{(MS-SSIM)}\\
			\midrule
			&Infusion training\cite{bordes2017learning}&&4.62$\pm$0.06&\\
			&ALI~\cite{dumoulin2016adversarially}(from\cite{warde2016improving})&&5.34$\pm$0.05&\\
			\parbox[t]{2mm}{\multirow{10}{*}{\rotatebox[origin=c]{90}{unsupervised}}}&Impr.GAN(-L+HA)\cite{salimans2016improved}&&6.86$\pm$0.06&\\
			&EGAN-Ent-VI~\cite{dai2017calibrating}     & &7.07$\pm$0.10&   \\
			&DFM~\cite{warde2016improving}     & &7.72$\pm$0.13&   \\
			&iWGAN~\cite{gulrajani2017improvedWGAN} & &7.86$\pm$0.07&   \\
			&iWGAN       & & 7.853$\pm$0.072&0.0504$\pm$0.0017\\
			&iWGAN-LC with AE clustering & 32 & 7.300$\pm$0.072&0.0507$\pm$0.0016\\
			&iWGAN-LC with RC clustering & 32 & 7.831$\pm$0.072&0.0491$\pm$0.0015\\
			&iWGAN-LC with RC clustering & 128 & 7.799$\pm$0.030&0.0491$\pm$0.0015\\
			&iWGAN-AC with AE clustering & 32 & 7.885$\pm$0.083&0.0504$\pm$0.0014\\
			&iWGAN-AC with RC clustering & 10 & 8.433$\pm$0.068&0.0505$\pm$0.0016\\
			&iWGAN-AC with RC clustering & 32 & \textbf{8.673$\pm$0.075}&	0.0500$\pm$0.0016\\
			&iWGAN-AC with RC clustering & 128 &\textbf{8.625$\pm$0.109}&\textbf{0.0465$\pm$0.0015}\\
			\midrule
			\parbox[t]{2mm}{\multirow{6}{*}{\rotatebox[origin=c]{90}{supervised}}}&iWGAN-LC &  & 7.710$\pm$0.084&0.0510$\pm$0.0013\\
			&Impr.GAN~\cite{salimans2016improved}&&8.09$\pm$0.07&\\
			&iWGAN-AC~\cite{gulrajani2017improvedWGAN} & &8.42$\pm$0.10&   \\
			&iWGAN-AC&&8.35$\pm$0.07&0.049$\pm$0.0018\\
			&AC-GAN~\cite{ACGAN}&&8.25$\pm$0.07&\\
			&SGAN~\cite{huang2016stacked}&&8.59$\pm$0.12&\\
			\midrule
			&CIFAR-10 (original data)&  &  11.237$\pm$0.116&0.0485$\pm$0.0016\\
			\bottomrule
		\end{tabular}
	}
	\caption{Comparison of Inception and diversity scores (lower score = higher diversity) on CIFAR-10. The unsupervised methods do not use the CIFAR-10 class labels. Note that our unsupervised methods achieve state-of-the-art performance comparable to the best supervised approaches.}
	\label{tab:inception_CIFAR-10}
\end{table}

\begin{table}[th]
	\centering
	\resizebox{\linewidth}{!}
	{
		\begin{tabular}{lr|c|c}
			\toprule
			\multirow{2}{*}{\textbf{Method}}     &\multirow{2}{*}{\textbf{Clusters}}  & \textbf{CORNIA} & \textbf{Diversity}\\
			&& \textbf{score} & \textbf{(MS-SSIM)}\\
			\midrule
			DCGAN-LC with AE clustering & 100 & 62.12$\pm$0.51&0.0475$\pm$0.0013\\
			iWGAN-LC with AE clustering & 100 & 60.24$\pm$0.61&0.0439$\pm$0.0010\\
			*iWGAN       & & 54.27$\pm$0.67&0.0488$\pm$0.0011\\
			*iWGAN-LC with RC clustering & 16 & 55.37$\pm$0.67&0.0490$\pm$0.0014\\
			*iWGAN-LC with RC clustering & 128 & 55.27$\pm$0.68&0.0484$\pm$0.0010\\
			\midrule
			LLD-icon (original data)&  &  61.00$\pm$0.62&0.0482$\pm$0.0014\\
			*LLD-icon-sharp (original data)&  &  55.37$\pm$0.67&0.0494$\pm$0.0011\\
			\bottomrule
		\end{tabular}
	}
	\caption{CORNIA scores and diversity scores for models trained on LLD-icon. The starred (*) models where trained on the subset LLD-icon-sharp. Lower values mean higher quality for CORNIA and higher diversity for MS-SSIM.}
	\label{tab:cornia}
	\vspace{-0.35cm}
\end{table}

\paragraph{Clustering}
On CIFAR-10, increasing the number of RC clusters from 1 to 128 leads to better diversity scores for iWGAN-AC, at the same time the Inception score peaks above 32 clusters. We note that using RC clustering leads to better performance than using AE clustering.

\paragraph{Performance and state-of-the-art}
Our best Inception score of $8.67$ achieved with iWGAN-AC and 32 RC clusters is significantly higher than $8.09$ by Salimans~\etal~\cite{salimans2016improved} with their Improved GAN method, the best score reported in the literature for unsupervised methods. Surprisingly, our best result, achieved with unsupervised synthetic labels provided by RC clustering, is comparable to $8.59$ of the Stacked GANs approach by Huang~\etal~\cite{huang2016stacked}, the best score reported for supervised methods.
\vspace{-0.2cm}
\paragraph{Image quality}
Complementary to the Inception and diversity scores we also measured the image quality using CORNIA, a robust no-reference image quality assessment method proposed by Ye and Doermann~\cite{CORNIA}. On both CIFAR-10 and LLD-icon our generative models obtained CORNIA scores equivalent to those of the original images from each dataset. This result is in-line with the findings in~\cite{ABC-GAN}, where the studied GANs also converge in terms of CORNIA scores towards the data image quality at GAN convergence. We show the CORNIA and MS-SSIM scores for the LLD-icon dataset, as a complement to the Inception scores on CIFAR-10, in Table~\ref{tab:cornia}.
\vspace{-0.2cm}
\paragraph{LC vs. AC for conditional GANs}
Our AC-GAN variants are better than their LC counterparts in terms of Inception scores, but comparable in terms of diversity for CIFAR-10. We believe that this is owed to fact that AC-GAN enforces the generation of images which can easily be classified to the provided clusters, which in turn could raise the classifier-based Inception score. Even though the numbers indicate a qualitative advantage of AC- over LC-GAN, we prefer the latter for our logo application as it allows smooth interpolations even in-between different clusters. This is not possible in the standard AC-GAN implementation since the cluster labels are discrete integer values and thus all our desirable latent space operations would be constrained to be performed within a specific data cluster, which does not match our intended use.

\section{Logo synthesis by latent space exploration}
\label{sec:logo_synthesis}

As mentioned in the previous section, layer conditioning allows for smooth transitions in the latent space from one class to another, which is critical for logo synthesis and manipulation by exploration of the latent space. Therefore, we work with two configurations for these experiments: iWGAN-LC with 128 RC clusters and DCGAN-LC with 100 AE clusters. Their Inception, diversity and CORNIA scores are comparable on the LLD-icon dataset.

\subsection{Sampling}
In generative models like GANs~\cite{goodfellow2014generative} and VAEs~\cite{kingma2013auto}, images are generated from a high-dimensional latent vector (with usually somewhere between 50 and 1000 dimensions), also commonly referred to as z-vector. During training, each component of this vector is randomly sampled from a Uniform or Gaussian distribution, so that the generator is trained to produce a reasonable output for any random vector sampled from the same distribution. The space spanned by these latent vectors, called the latent space, is often highly structured, such that latent vectors can be deliberately manipulated in order to achieve certain properties in the output~\cite{brock2016neural, dosovitskiy2015learning, radford2015unsupervised}. 

\begin{figure*}[thbp!]
\centering
\resizebox{\linewidth}{!}
{
\begin{tabular}{cccc}
	\includegraphics[width=0.24\textwidth]{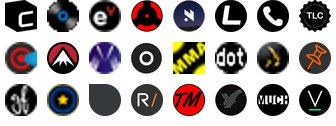} &
	\includegraphics[width=0.24\textwidth]{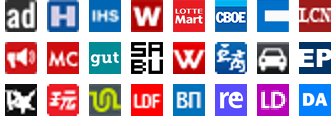} &
	\includegraphics[width=0.24\textwidth]{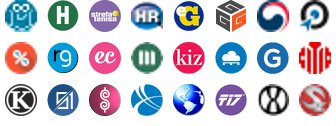} &
	\includegraphics[width=0.24\textwidth]{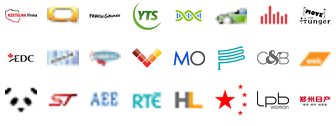} \\
	\includegraphics[width=0.24\textwidth]{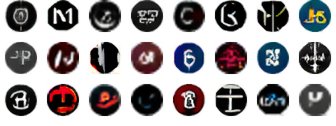} &
	\includegraphics[width=0.24\textwidth]{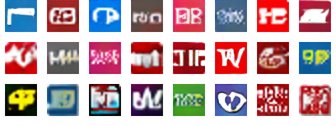} &
	\includegraphics[width=0.24\textwidth]{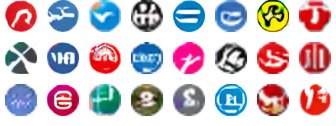} &
	\includegraphics[width=0.24\textwidth]{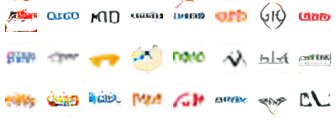}
\end{tabular}
}
\vspace{-0.3cm}
	\caption{The first four (random) clusters of LLD-icon as attained with our AE-Clustering method using 100 cluster centers. The top half of each example contains a random selection of original images, while the bottom half consists of samples generated by DCGAN-LC for the corresponding cluster. The very strong visual correspondence demonstrates the network's ability to capture the data distributions inherent the classes produced by our clustering method.}
	\label{fig:clusters}
    \vspace{-0.3cm}
\end{figure*}

Using DCGAN-LC with 100 AE clusters on the same data, Figure~\ref{fig:clusters} contains samples from a specific cluster next to a sample of the respective original data. This shows how the layer conditional DCGAN is able to pick up on the data distribution and produce samples which are very easy to attribute to the corresponding cluster and are often hard to distinguish from the originals at first glance.
For comparison we also show results for iWGAN-LC with 128 RC clusters trained on the LLD-icon-sharp dataset in Figure~\ref{fig:teaser}.

\subsection{Interpolations}
To show that a generator does not simply learn to reproduce samples from the training set, but is in fact able to produce smooth variations of its output images, it is common practice~\cite{goodfellow2016nips} to perform interpolations between two points in the latent space and to show that the outcome is a smooth transition between the two corresponding generated images, with all intermediate images exhibiting the same distribution and quality. Interpolation also provides an effective tool for a logo generator application, as the output image can be manipulated in a controlled manner towards a certain (semantically meaningful) direction in latent space.

\begin{figure}[t!]
\centering
\includegraphics[width=0.8\linewidth]{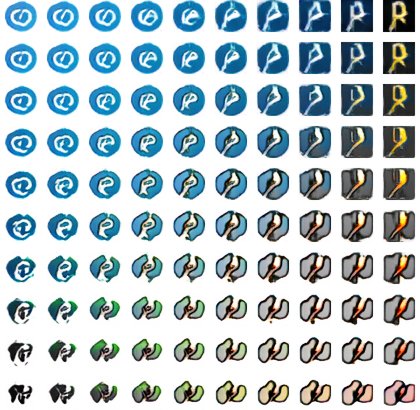}
   \caption{Interpolation between 4 selected logos of distinct classes using DCGAN-LC with 100 AE clusters on LLD-icon, showcasing smooth transitions and interesting intermediate samples in-between all of them.} 
\label{fig:interp_large}
\vspace{-0.3cm}
\end{figure}

For all our interpolation experiments we use the distribution matching methods from~\cite{agustsson2017optimal} in order to preserve the prior distribution the sampled model was trained on. An example with 64 interpolation steps to showcase the smoothness of such an interpolation is given in Figure~\ref{fig:interp_large} where we interpolate between 4 sample points, producing believable logos at every step. As it is the case in this example, the interpolation works very well even between logos of different clusters, even though the generator was never trained for mixed cluster attributes.

\begin{figure}[tbp]
\centering
\begin{tabular}{c}
\includegraphics[width=0.8\linewidth]{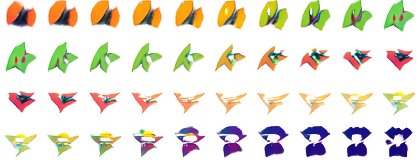}\\
\midrule
\vspace{-0.1cm}
\\
\includegraphics[width=0.8\linewidth]{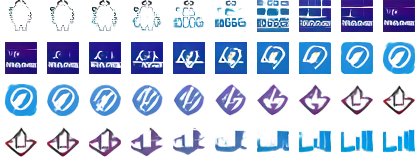}\\
\end{tabular}
\vspace{-0.3cm}
   \caption{Continuous interpolation between 5 random points each within one cluster (top) and in-between distinct clusters (bottom) in latent space using iWGAN-LC with 128 RC clusters on icon-sharp. We observe smooth transitions and logo-like samples in all of the sampled subspace.}
\label{fig:interp8}
\vspace{-0.3cm}
\end{figure}
Some more interpolations between different logos both within a single cluster and between logos of different clusters are shown in Figure~\ref{fig:interp8}, this time between 2 endpoints and with only 8 interpolation steps.

\subsection{Class transfer}
\begin{figure}[tb]
\centering
\includegraphics[width=1\linewidth]{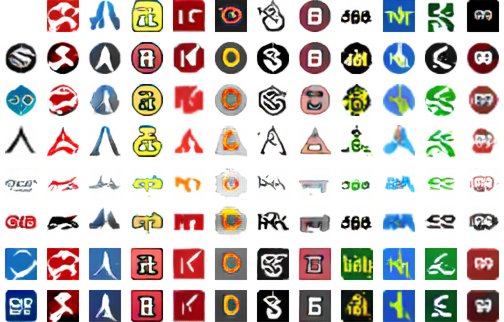}
\vspace{-0.3cm}
   \caption{Logo class transfer using DCGAN-LC on LLD-icon with 100 AE clusters. The logos of the 1st row get transferred to the class (cluster) of the logos in the 1st column (to the left). Hereby the latent vector is kept constant within each column and the class label is kept constant within each row (except for the 1st ones, resp.). The original samples have been hand-picked for illustrative purposes.}
\label{fig:class_transfer}
\end{figure}

As the one-hot class vector representing the logo cluster is separate from our latent vector, it is also possible to keep the latent space representation constant and only change the cluster of a generated logo. Figure~\ref{fig:class_transfer} contains 11 logos (top row) that are being transformed to a particular cluster class in each subsequent row. This shows how the general appearance such as color and contents are encoded in the z-vector while the cluster label transforms these attributes into a form that conforms with the contents of the respective cluster. Here, again, interpolation could be used to create intermediate versions as desired.

\subsection{Vicinity sampling}
\begin{figure}[tb]
\centering
\includegraphics[width=0.65\linewidth]{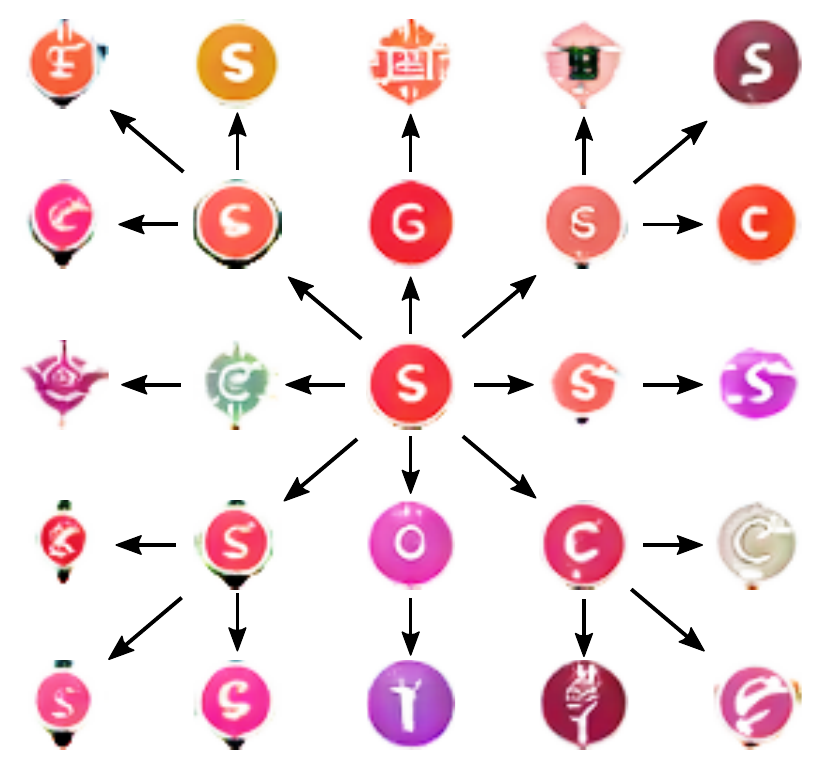}
   \caption{Vicinity Sampling using iWGAN-LC on LLD-icon-sharp with 128 RC clusters.}
\label{fig:vic_sampling}
\end{figure}
Another powerful tool to explore the latent space is vicinity sampling, where we perturb a given sample in random directions of the latent space. This could be useful to present the user of a logo generator application with a choice of possible variants, allowing him to modify his logo step by step into directions of his choice. In Figure~\ref{fig:vic_sampling} we present an example of a 2-step vicinity sampling process, where we interpolate one-third towards random samples to produce a succession of logo variants.

\subsection{Vector arithmetic example: Sharpening}
\begin{figure}[htb]
	\centering
	\hfill
    \subfloat[Original samples]{
    	\includegraphics[width=0.3\linewidth]{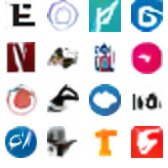}}
    \hfill
	\subfloat[Sharpened samples]{
		\includegraphics[width=0.3\linewidth]{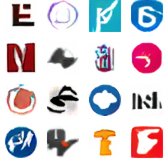}}
	\hfill
	\hfill
    \vspace{-0.1cm}
	\caption{Sharpening of logos in the latent space by adding an offset calculated from the latent vectors of sharp and blurry samples. We used DCGAN-LC and 100
AE clusters.}
	\label{fig:sharpening}
\end{figure}

For models trained on our LLD-icon data, some of the generated icons are blurry since roughly half of the logos in this dataset are upscaled from a lower resolution. However, by averaging over the z-vector of a number of blurry samples and subtracting from this the average of a number of sharp samples, it is possible to construct a ``sharpening'' vector which can be added to blurry logos to transform them into sharp ones. This works very well even if the directional vector is calculated exclusively from samples in one cluster and then applied samples of another, showing that the blurriness is in fact nothing more than a feature embedded in latent space. The result of such a transformation is shown in Figure~\ref{fig:sharpening}, where such a sharpening vector was calculated from 40 sharp and 42 blurry samples manually selected from two random batches of the same cluster. The resulting vector is then applied equally to all blurry samples. The quality of the result, while already visually convincing, could be further optimized by adding individually adjusted fractions of this sharpening vector to each logo.

This example of adding a sharpening vector to the latent representation is only one of many latent space operations one could think of, such as directed manipulation of form and color as performed in the supplementary material.

\section{Conclusions}
\label{sec:conclusions}
In this paper we tackled the problem of logo design by synthesis and manipulation with generative models:
\begin{itemize}
\setlength{\itemsep}{0pt} 
\setlength{\parskip}{0pt}
\item[\textbf{(i)}] We introduced a Large Logo Dataset (LLD) crawled from Internet with orders of magnitude more logos than the existing datasets. 
\item[\textbf{(ii)}] In order to cope with the high multi-modality and to stabilize GAN training on such data we proposed clustered GANs, that is GANs conditioned with synthetic labels obtained through clustering. We performed clustering in the latent space of an Autoencoder or in the CNN features space of a ResNet classifier and conditioned DCGAN and improved WGAN utilizing either an Auxiliary Classifier or Layer Conditional model. 
\item[\textbf{(iii)}] We quantitatively validated our clustered GAN approaches on a CIFAR-10 benchmark where we set a clear state-of-the-art Inception score for unsupervised generative models, showcasing the benefits of meaningful synthetic labels obtained through clustering in the CNN feature space of a an ImageNet classifier.
\item[\textbf{(iv)}] We showed that the latent space of the networks trained on our logo data is smooth and highly structured, thus having interesting properties exploitable by performing vector arithmetic in that space. 
\item[\textbf{(v)}] We showed that the synthesis and manipulation of (virtually) infinitely many variations of logos is possible through latent space exploration equipped with a number of operations such as interpolations, sampling, class transfer or vector arithmetic in latent space like our sharpening example.
\end{itemize}
Our solutions ease the logo design task in an interactive manner and are significant steps towards a fully automatic logo design system.

For more results, operations, and settings the reader is invited to consult the supplementary material.

{\small
	\bibliographystyle{ieee}
	\bibliography{bibliography}
}

\appendix

\section*{Supplementary Material}
The following pages contain the supplementary material for this paper.

After presenting some latent space exploration experiments with LLD-logo in Section~\ref{sec:vecarim}, we give some additional details on the data collection process as well as the final contents of our LLD datasets in Section~\ref{sec:lld-stats}. We then proceed to show in Section~\ref{sec:samples}, for each subset of our Large Logo Dataset, an excerpt of the collected data together with generated samples from selected GAN architectures and the clusters produced by the applied clustering methods. For CIFAR-10 we also show samples from our cluster-conditional CIFAR-10 models together with samples from the unconditional and supervised iWGAN variants in this section. Finally, we give some details on architecture and training hyper-parameters of our models in Section~\ref{sec:architecture}..

\clearpage
\pagebreak
\section{Latent space exploration on LLD-logo}
\label{sec:vecarim}
In this section, we present some interpolations on the LLD-logo dataset and perform two additional experiments with latent space operations.

\begin{figure}[htbp]
	\centering
	\begin{subfloat}[Interpolation between 4 square logos.]{
			\centering
			\includegraphics[width=1.1\linewidth]{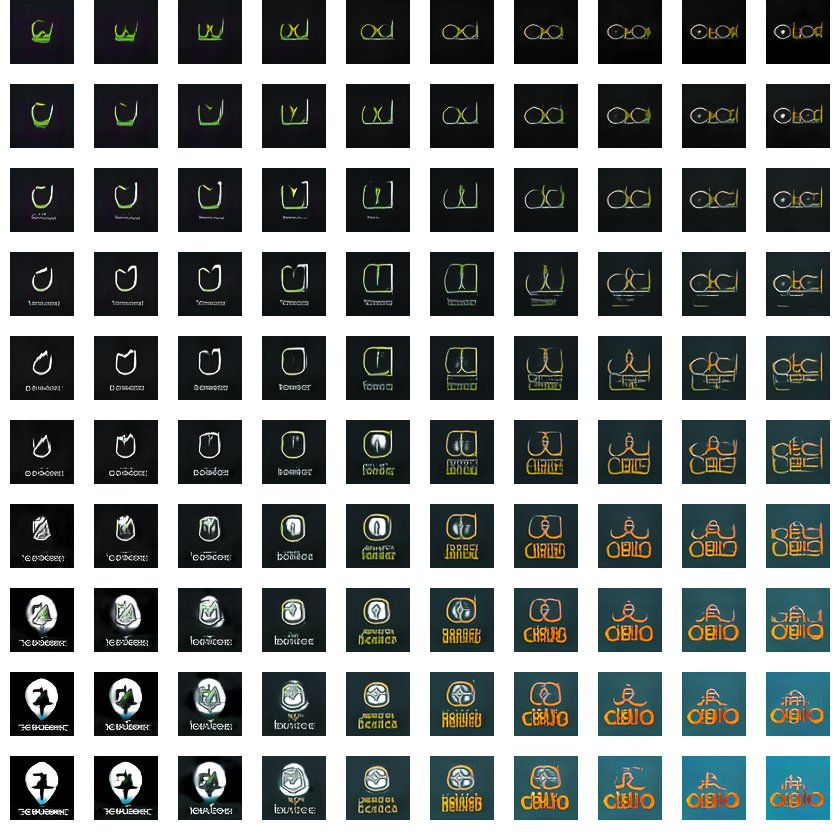}
		}
	\end{subfloat}
	\begin{subfloat}[Interpolation between logos of different shape.]{
			\centering
			\includegraphics[width=1.1\linewidth]{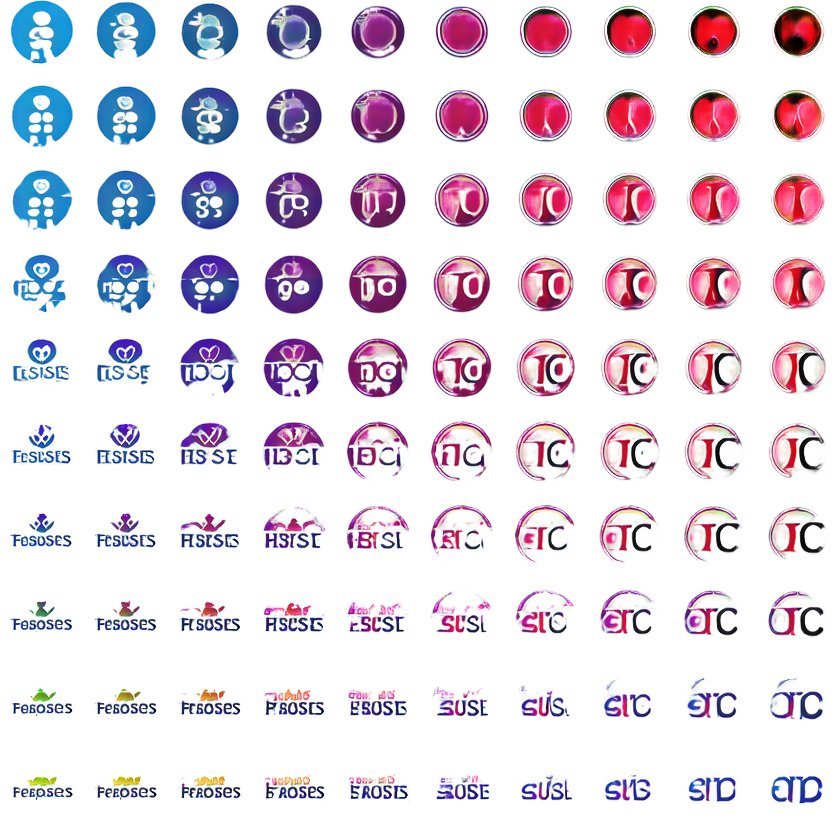}
		}
	\end{subfloat}
	\caption{Four-point interpolation on LLD-logo.}
	\label{fig:4p-twitter}
\end{figure}

\paragraph{Interpolation}
In Figure~\ref{fig:4p-twitter} we present two examples of interpolations between 4 different samples, representing a small section of the high-dimensional logo manifold created by the GAN.

\begin{figure}[tbp]
	\centering
	\vspace{-1cm}
	\begin{subfloat}[Random Sample, unmodified.]{
			\label{fig:ar-orig-a}
			\centering
			\includegraphics[width=0.85\linewidth]{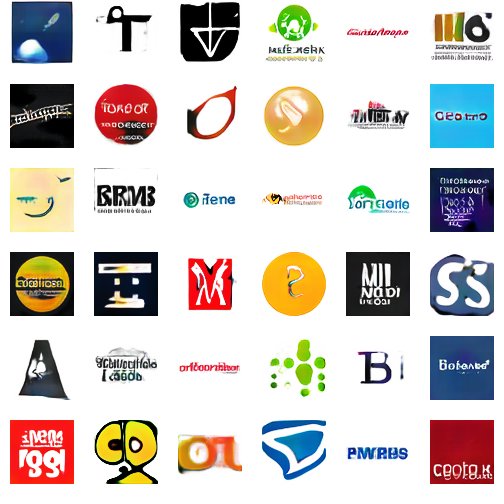}
		}
	\end{subfloat}
	\begin{subfloat}[Sample from \protect\subref{fig:ar-orig-a} shifted towards blue logos]{
			\centering
			\includegraphics[width=0.85\linewidth]{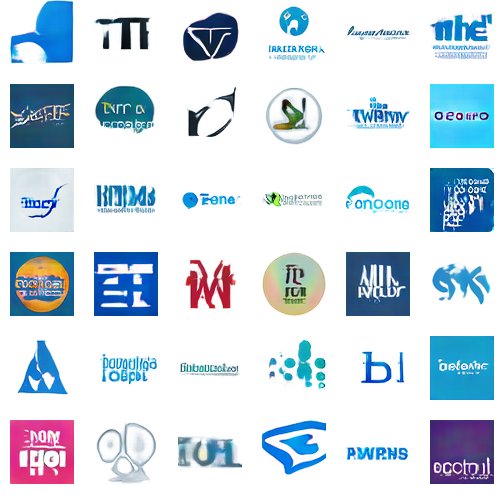}
		}
	\end{subfloat}
	\begin{subfloat}[Sample from \protect\subref{fig:ar-orig-a} shifted towards red logos]{
			\centering
			\includegraphics[width=0.85\linewidth]{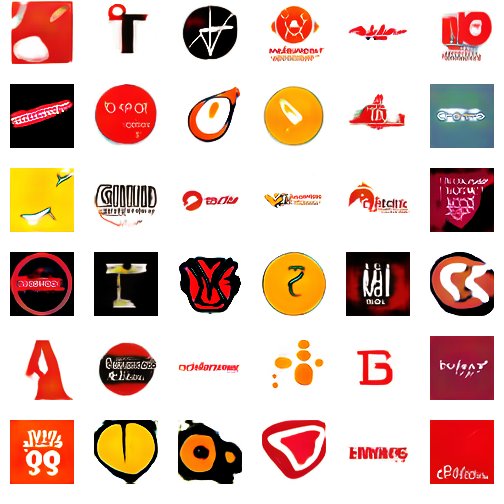}
		}
	\end{subfloat}
	\caption{Blue-red shift on a random batch. Directional vectors are both applied in latent space and in cluster label space.}
	\label{fig:ar-orig}
\end{figure}

\begin{figure}[tbp]
	\centering
	\begin{subfloat}[Samples from Figure~\ref{fig:ar-orig-a} shifted towards red logos only in latent vector space]{
			\centering
			\includegraphics[width=0.85\linewidth]{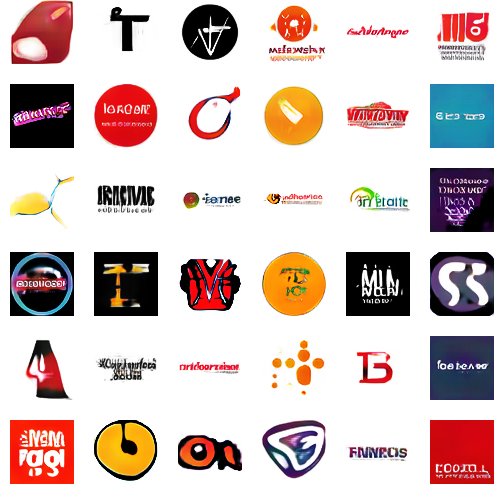}
		}
	\end{subfloat}
	\begin{subfloat}[Samples from Figure~\ref{fig:ar-orig-a} shifted towards red logos only in label vector space]{
			\centering
			\includegraphics[width=0.85\linewidth]{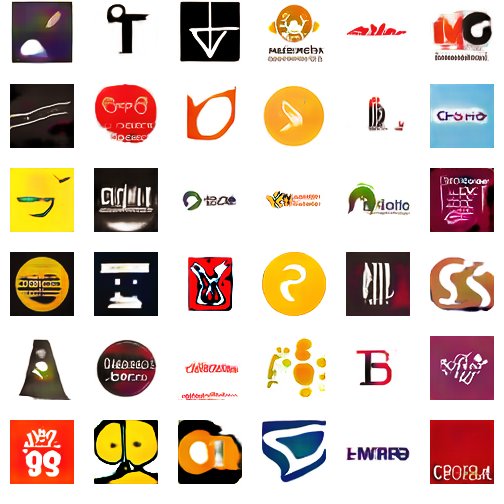}
		}
	\end{subfloat}
	\caption{Blue-red shift on a random batch performed in either latent representation or cluster labels.}
	\label{fig:ar-colzy}
\end{figure}

\paragraph{Vector arithmetic}
First, we define two desirable operations we would like to perform (1) Color shifts from red to blue and blue to red and (2) Shape changes from square to round and round to square.
For each of those semantic operations we identify a number (for our experiments around 30) of samples that match our criteria. To get operation (1) this means we select 30 red and 30 blue logos. We then construct a directional vector by subtracting the mean latent space vector of all blue logos from the mean latent space vector of all red logos, which gives us a directional vector from red to blue. Since some of these semantic attributes are expected to be encoded in the cluster labels as well, we can do the same with our one-hot encoded class vectors, which we can view as an additional cluster space.
In Figure~\ref{fig:ar-orig} we add this directional vector to a new random batch of generated logos. If we subtract the directional vector, we get a shift in the opposite direction, i.e. from blue to red. To find out how much of the color information is encoded in the latent representation and in the clusters respectively, we can perform the operation in only one of these domains. This is done in Figure~\ref{fig:ar-colzy} for the red-shift, where we observe a very similar behavior for both spaces, indicating that the color information is equally encoded in both latent space and labels.

\begin{figure}[tbp]
	\vspace{-1cm}
	\centering
	\begin{subfloat}[Random Sample, unmodified. (Same as Figure~\protect\ref{fig:ar-orig-a})]{
			\label{fig:ar-orig-shape}
			\centering
			\includegraphics[width=0.85\linewidth]{images/vecarim_orig.jpg}
		}
	\end{subfloat}
	\begin{subfloat}[Samples from Figure~\protect\subref{fig:ar-orig-shape} shifted towards round logos.]{
			\centering
			\includegraphics[width=0.85\linewidth]{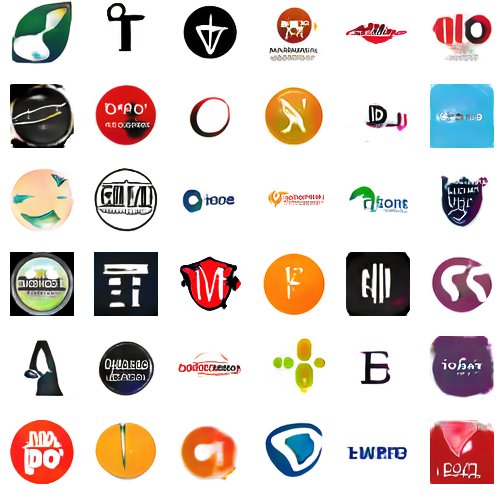}
		}
	\end{subfloat}
	\begin{subfloat}[Samples from Figure~\protect\subref{fig:ar-orig-shape} shifted towards square logos]{
			\centering
			\includegraphics[width=0.85\linewidth]{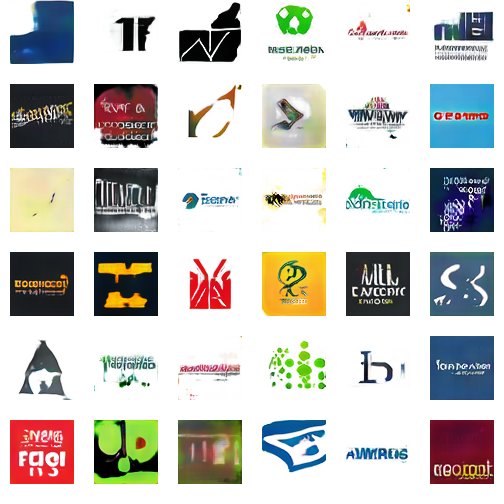}
		}
	\end{subfloat}
	\caption{Round-square shape shift on a random batch. Directional vectors are both applied in latent space and in cluster label space.}
	\label{fig:ar-shape}
\end{figure}

\begin{figure}[tbp]
	\centering
	\begin{subfloat}[Samples from Figure~\protect\ref{fig:ar-orig-shape} shifted towards red logos only in latent vector space]{
			\centering
			\includegraphics[width=0.85\linewidth]{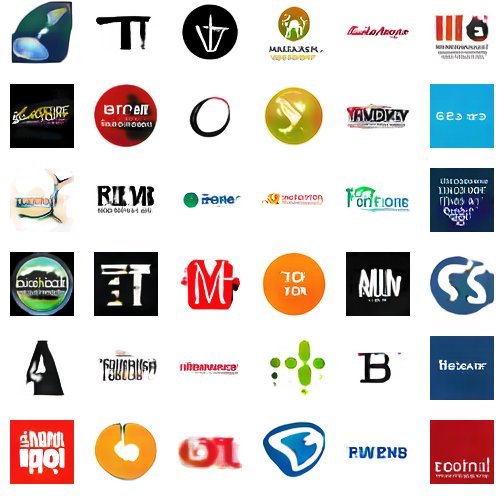}
		}
	\end{subfloat}
	\begin{subfloat}[Samples from Figure~\protect\ref{fig:ar-orig-shape} shifted towards red logos only in label vector space]{
			\centering
			\includegraphics[width=0.85\linewidth]{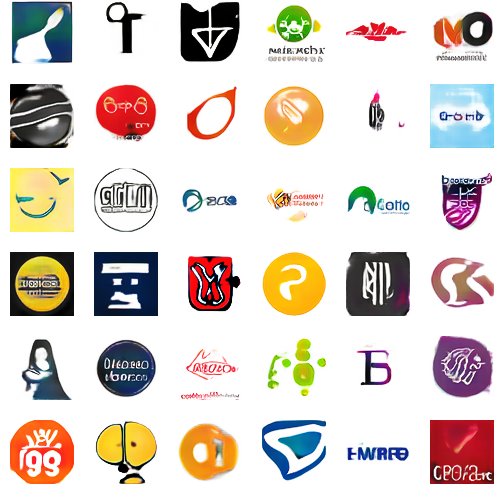}
		}
	\end{subfloat}
	\caption{Round-square shape shift on a random batch performed in either latent representation or cluster labels.}
	\label{fig:ar-shapexy}
\end{figure}

Our second experiment is performed in the same way, and the directional vector is applied to the same batch of samples. Figure~\ref{fig:ar-shape}, again, shows the result for a simultaneous addition of both (latent and class) vectors in each direction, whereas each space is considered individually in Figure~\ref{fig:ar-shapexy} for the directional vectors towards round logos. Here we can observe that some logos respond better to the change in latent space, while others seem more responsive to a changing cluster label. Overall, the label information seems to be a little stronger in this case. 

In both experiments, the combined shift clearly performs best, and could provide a powerful tools for logo manipulation and other applications.

\clearpage
\pagebreak

\clearpage
\pagebreak
\section{LLD crawling and image statistics}
\label{sec:lld-stats}
\subsection{LLD-icon}

When collecting the favicons for LLD-icon, our download script directly converted all icons found to a standardized 32x32 pixel resolution and RGB color space, discarding all non-square images.
After acquiring the raw data from the web, we remove all exact duplicates and perform a three-stage clean-up process:
\begin{enumerate}[noitemsep,nolistsep]
	\item Sort all images by complexity by evaluating its PNG-compressed file size
	\item Manually inspect and partition the resulting sorted list into three sections: Clean, mostly clean and mostly unwanted data. The last section is discarded, while the middle part (mostly clean) is further processed in the next step.
	\item Sort the intermediate section according to the number of white pixels in each image and cut off at a certain point after inspection, discarding the images containing the least amount of white pixels.
\end{enumerate}
Table~\ref{tab:icon-crawling} shows statistics on the crawling process, original image resolutions the icons where rescaled from, and numbers on content removed through our clean-up process.

\subsection{LLD-logo}
\begin{figure}[t]
	\centering
	\includegraphics[width=0.8\linewidth]{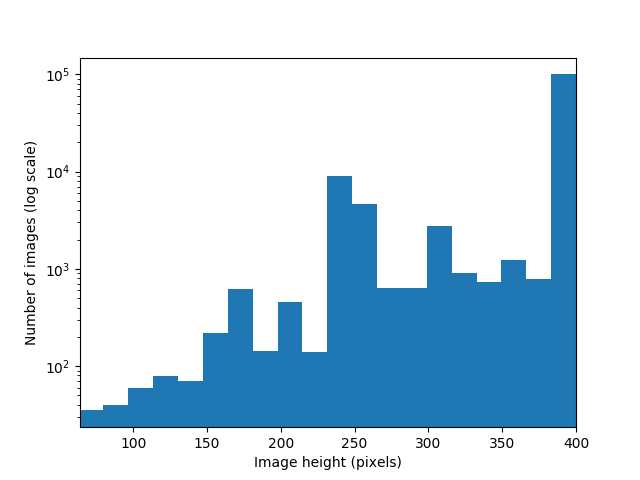}
	\caption{Histogram of image sizes in LLD-logo. There are a total of 329 different image resolutions contained in the dataset.}
	\label{fig:twitter_hist}
\end{figure}

During the collection of LLD-logo on twitter, we use a face detector recognize faces and proceed to the next user in the search results if a face was detected. At the same time, we make use of twitters (relatively new) sensitive content flag to reject such flagged profiles. 
As the number of rejected profiles in Table~\ref{tab:twitter} compared to the number of discarded images during cleanup (of which a substantial number where due to sensitive content) shows, this flag is only used very sporadically at this time, and is far from a reliable indicator.
Figure~\ref{fig:twitter_hist} shows a histogram of image resolutions contained in LLD-icon (where no re-scaling was performed during data collection), with the top-5 image resolutions (amounting to 92\% of images) given in Table~\ref{tab:twitter-top-5}.
\vspace{-0.3cm}
\vspace{-0.2cm}
\begin{table}[!ht]
	\begin{center}
		\begin{tabular}{lrl}
			\toprule
			Failed requests & $150,413$ & \\
			Unreadable files & $71,596$ & \\
			Non-square images & $36,401$ & \\
			Unable to process & $6$ & \\
			Total images saved & $662,273$ & \\
			\midrule
			\multicolumn{3}{c}{Image re-scaling} \\
			Native 32 p & $158,881$ & $24.0 \%$\\
			Scaled up & $355,260$ & $53.6 \%$\\
			Scaled down  & $148,132$ & $22.4 \%$\\
			\midrule
			\multicolumn{3}{c}{Dataset cleanup} \\
			Duplicates removed & $114,063$ & $17.2 \%$\\
			Discarded due to content & $61,833$ & \\
			Clean dataset size & $486,377$ & \\
			\bottomrule
		\end{tabular}
	\end{center}
	\vspace{-0.3cm}
	\caption{Crawling statistics for LLD-icon}
	\label{tab:icon-crawling}
\end{table}
\vspace{-0.2cm}
\begin{table}[t]
	\begin{center}
		\begin{tabular}{lr}
			\toprule
			Flagged content ignored & $1,066$\\
			Downloaded images & $182,998$\\
			Discarded during cleanup & $60,078$\\
			Final dataset size & $122,920$\\
			\bottomrule
		\end{tabular}
	\end{center}
	\vspace{-0.3cm}
	\caption{Crawling and clean-up statistics for LLD-icon}
	\label{tab:twitter}
\end{table}
\vspace{-0.2cm}
\begin{table}[t]
	\begin{center}
		\begin{tabular}{crr}
			\toprule
			Image height (px) & Number of images & $\%$ of total\\
			\midrule
			$400$ & $98,824$ & $80.4 \%$\\
			$240$ & $8,625$ & $7.0 \%$\\
			$256$ & $2,498$ & $2.0 \%$\\
			$300$ & $2,143$ & $1.7 \%$\\
			$250$ & $1,502$ & $1.2 \%$\\
			\bottomrule
		\end{tabular}
	\end{center}
	\vspace{-0.3cm}
	\caption{The 5 most prominent image resolutions in LLD-logo, covering 92.3\% of the contained images.}
	\label{tab:twitter-top-5}
\end{table}

\clearpage
\pagebreak
\section{Logo Data, clusters and generated samples}
\label{sec:samples}
In this section, we will show a small sample from each of our introduced datasets and present generated icons from models trained on said dataset. Additionally, we show the data clusters produced by our clustering methods.

\begin{figure*}[tbp]
	\centering
	\begin{subfloat}[Original data]{
			\centering
			\includegraphics[width=\textwidth]{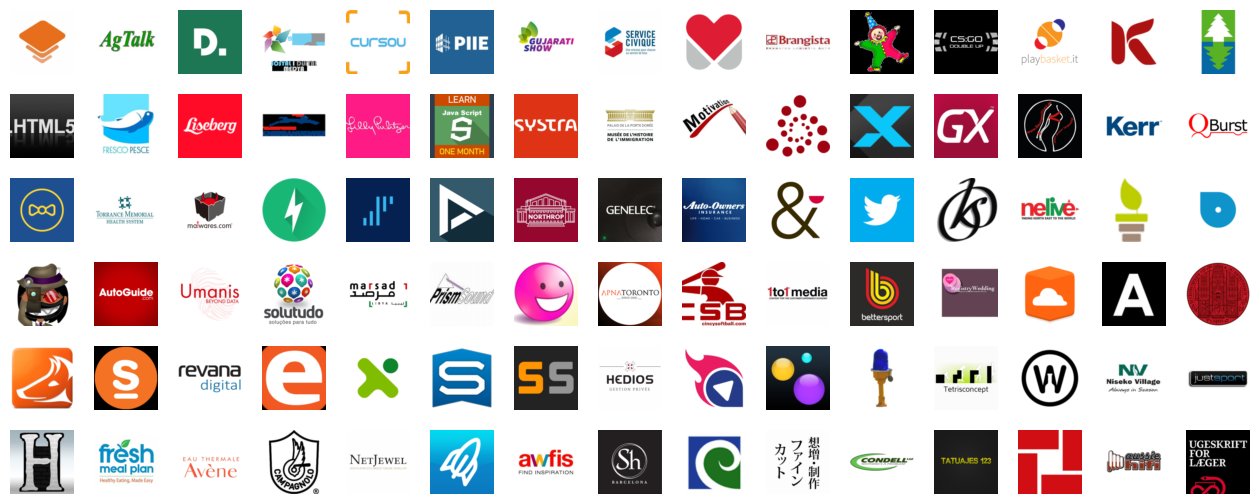}
		}
	\end{subfloat}
	\begin{subfloat}[iWGAN-LC with 64 RC clusters]{
			\centering
			\includegraphics[width=\textwidth]{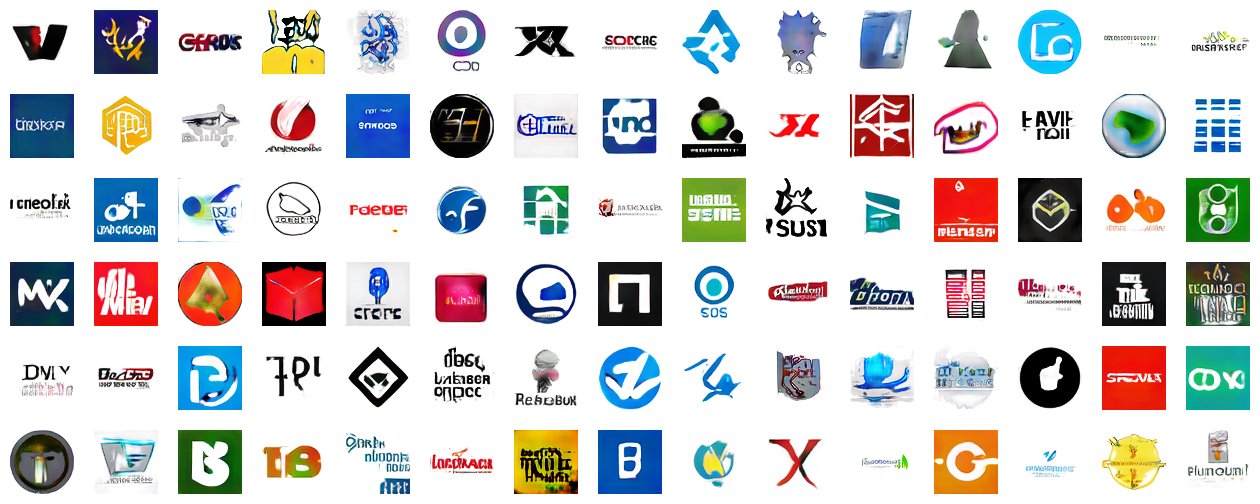}
		}
	\end{subfloat}
	\caption{Random samples from LLD-logo data and trained iWGAN model using 64 RC clusters and a 64\x64 pixel output resolution.}
	\label{fig:LLD-twitter}
\end{figure*}

\begin{figure*}[tbp]
	\centering
	\vspace{-0.5cm}
	\includegraphics[width=\textwidth]{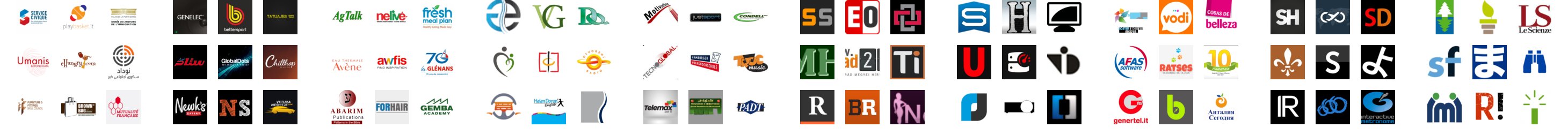}
	\vspace{0.2cm}
	\includegraphics[width=\textwidth]{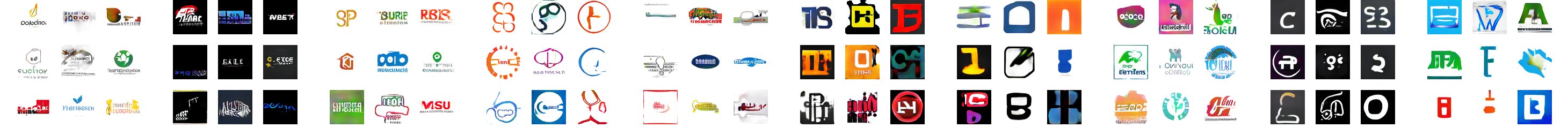}
	\includegraphics[width=\textwidth]{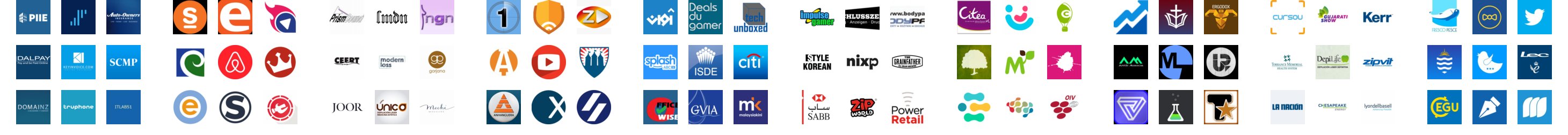}
	\vspace{0.2cm}
	\includegraphics[width=\textwidth]{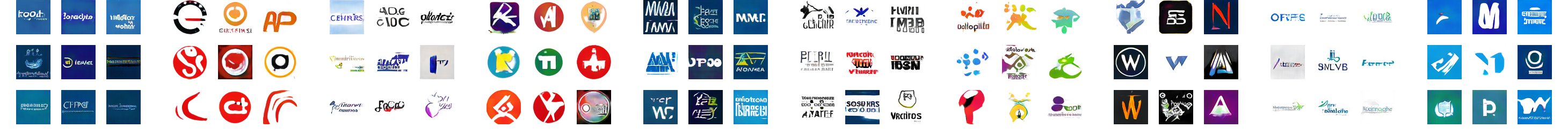}
	\includegraphics[width=\textwidth]{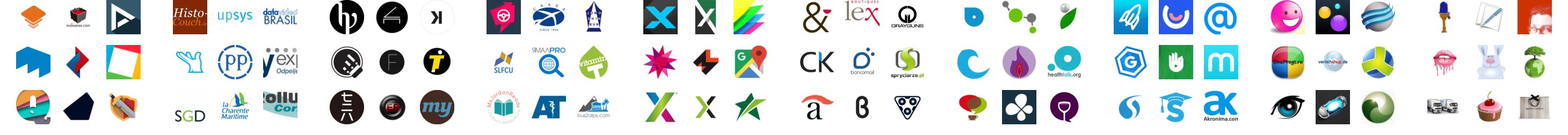}
	\vspace{0.2cm}
	\includegraphics[width=\textwidth]{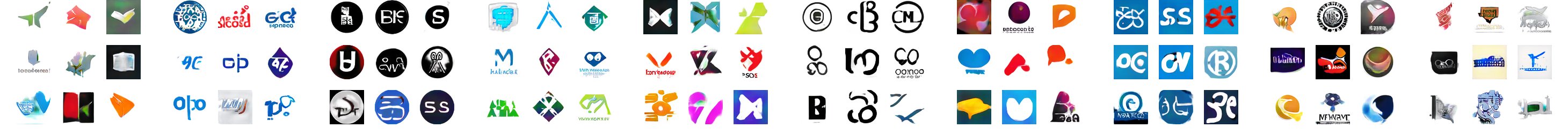}
	\includegraphics[width=\textwidth]{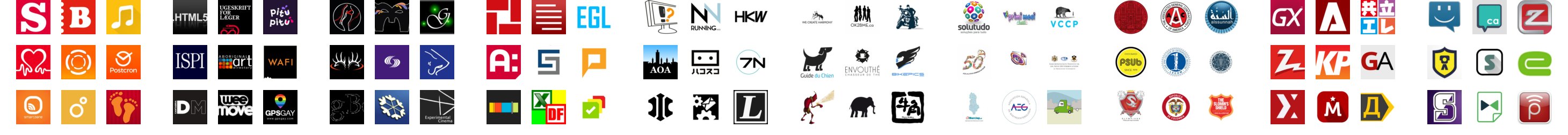}
	\vspace{0.2cm}
	\includegraphics[width=\textwidth]{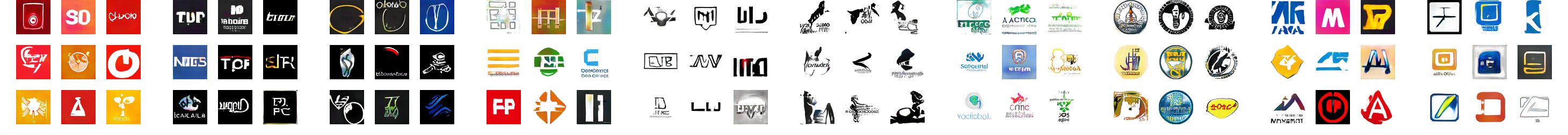}
	\includegraphics[width=\textwidth]{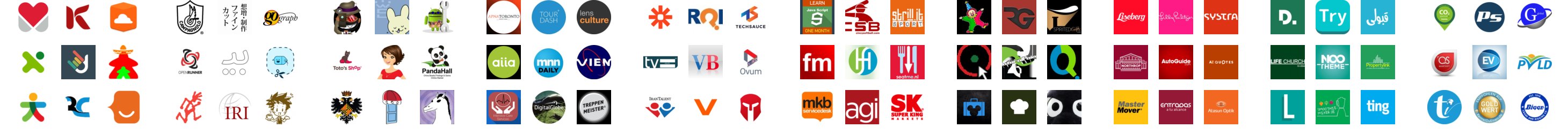}
	\vspace{0.2cm}
	\includegraphics[width=\textwidth]{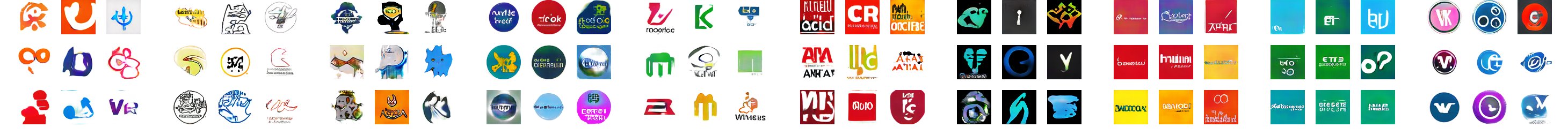}
	\includegraphics[width=\textwidth]{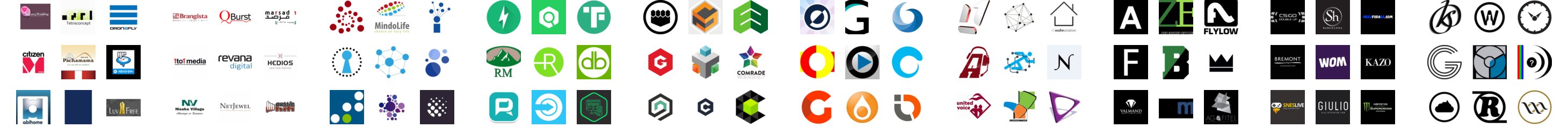}
	\vspace{0.2cm}
	\includegraphics[width=\textwidth]{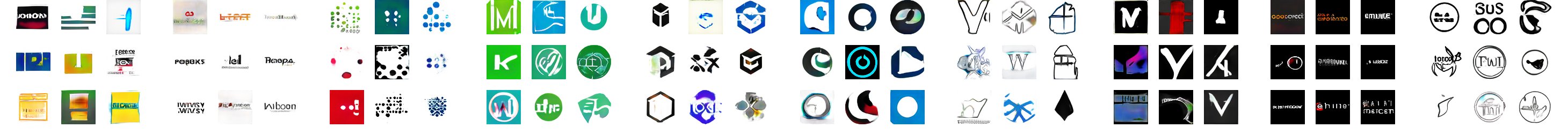}
	\includegraphics[width=\textwidth]{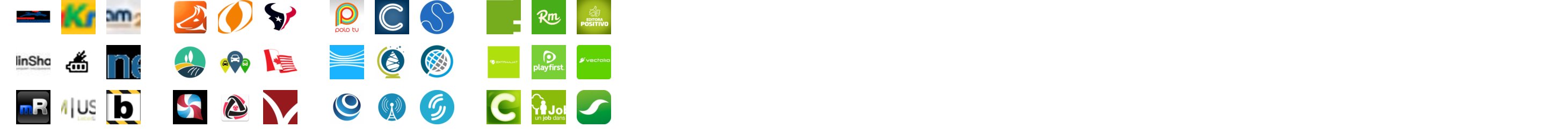}
	\vspace{0.2cm}
	\includegraphics[width=\textwidth]{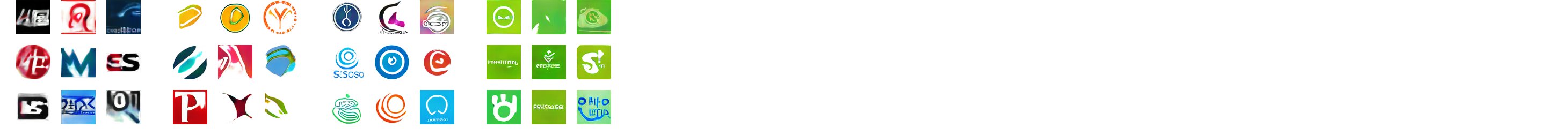}
	\vspace{-0.5cm}
	\caption{All 64 clusters of LLD-logo clustered with a ResNet classifier for 64 cluster centers. The top half of each block contains 9 random samples of original images from the cluster, while the bottom half contains 9 random samples from the iWGAN-LC Generator trained at 64\x64 pixels. Best viewed as PDF at 400\% magnification.}
	\label{fig:LLD-logo-cl}
\end{figure*}

Starting with LLD-logo, Figure~\ref{fig:LLD-twitter} shows a sample of the original data collected (reduced to 64\x64 pixels) next to the logos generated by an iWGAN model trained at 64\x64 pixels. Compared to LLD-icon, these logos contain a lot more text and sometimes more detailed images. Both of these features are recreated nicely by the model, where the text is often (but not always) illegible while still of a realistic appearance. We expect the legibility of the text to be much higher if our data would not contain a lot of non-latin (e.g. Chinese) characters. Figure~\ref{fig:LLD-logo-cl} contains the 64 clusters found by clustering with our RC method, showing very obvious semantic similarities within each cluster. It is not immediately noticeable that each block is composed of real (top half) and generated (bottom half) samples, which shows how well the GAN is able to reproduce the specific distributions inherent in each cluster.

\begin{figure*}[tbp]
	\centering
	\begin{subfloat}[Original data]{
			\centering
			\includegraphics[width=\textwidth]{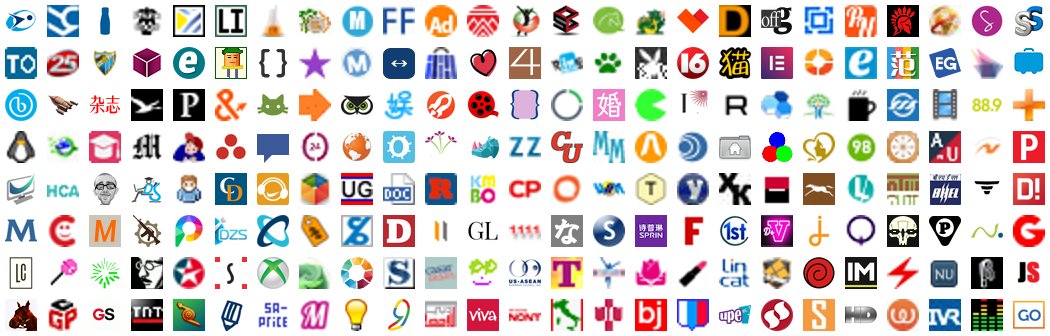}
		}
	\end{subfloat}
	\begin{subfloat}[DCGAN-LC with 100 AE clusters]{
			\centering
			\includegraphics[width=\textwidth]{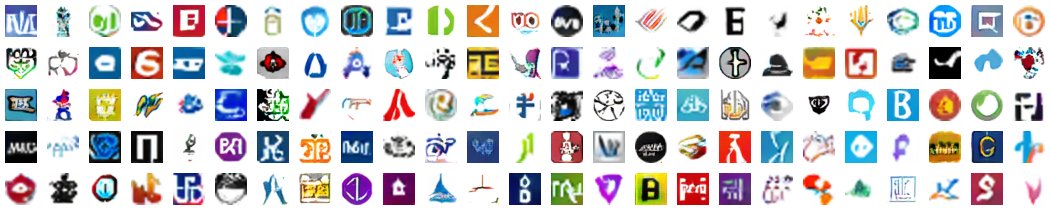}
		}
	\end{subfloat}
	\begin{subfloat}[iWGAN-LC with 100 AE clusters]{
			\centering
			\includegraphics[width=\textwidth]{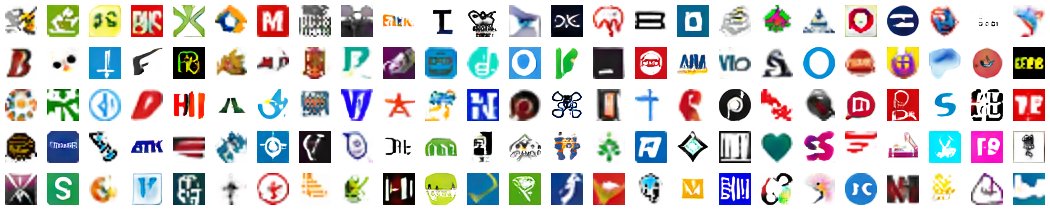}
		}
	\end{subfloat}
	\begin{subfloat}[iWGAN-LC with 128 RC Clusters]{
			\centering
			\includegraphics[width=\textwidth]{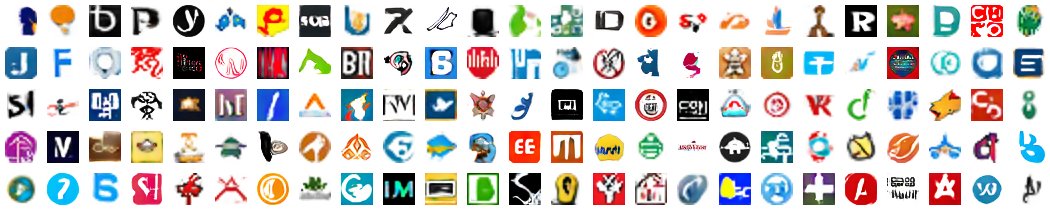}
		}
	\end{subfloat}
	\caption{Random samples from LLD-icon and generative models trained on this data.}
	\label{fig:LLD-icon-clean}
\end{figure*}

\begin{figure*}[tbp]
	\centering
	\begin{subfloat}[Original data]{
			\centering
			\includegraphics[width=\textwidth]{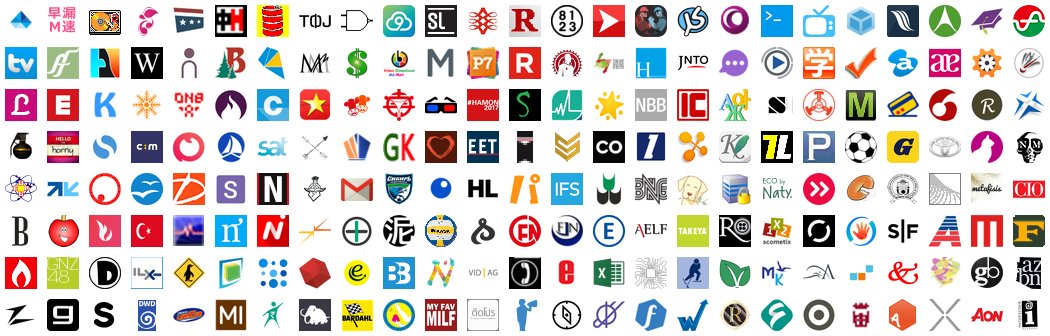}
		}
	\end{subfloat}
	\begin{subfloat}[Unconditional iWGAN]{
			\centering
			\includegraphics[width=\textwidth]{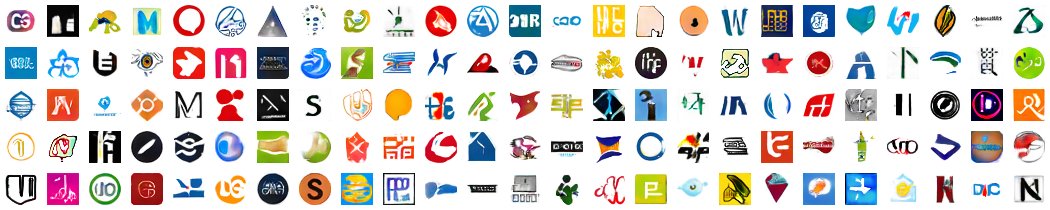}
		}
	\end{subfloat}
	\begin{subfloat}[iWGAN-LC with 16 RC clusters]{
			\centering
			\includegraphics[width=\textwidth]{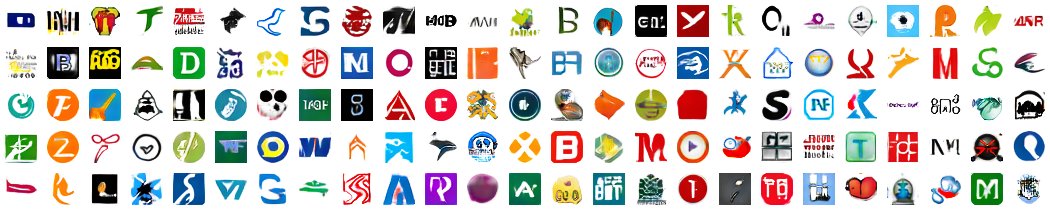}
		}
	\end{subfloat}
	\begin{subfloat}[iWGAN-LC with 128 RC Clusters]{
			\centering
			\includegraphics[width=\textwidth]{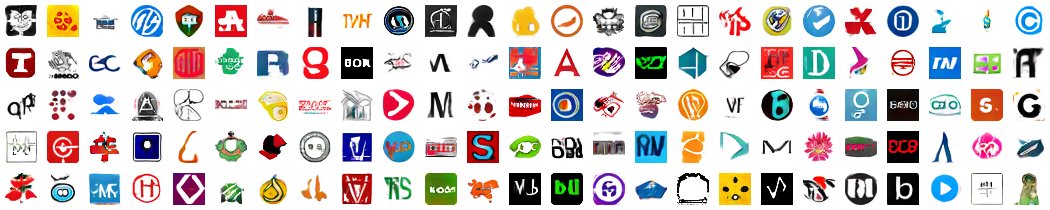}
		}
	\end{subfloat}
	\caption{Random samples from LLD-icon-sharp and generative models trained on this data.}
	\label{fig:LLD-icon-sharp}
\end{figure*}

\begin{figure*}[tbp]
	\centering
	\vspace{-0.5cm}
	\includegraphics[width=\textwidth]{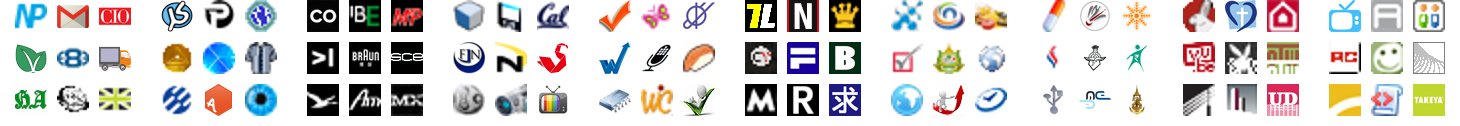}
	\vspace{0.2cm}
	\includegraphics[width=\textwidth]{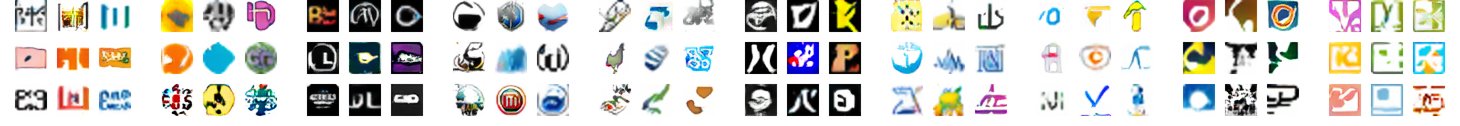}
	\includegraphics[width=\textwidth]{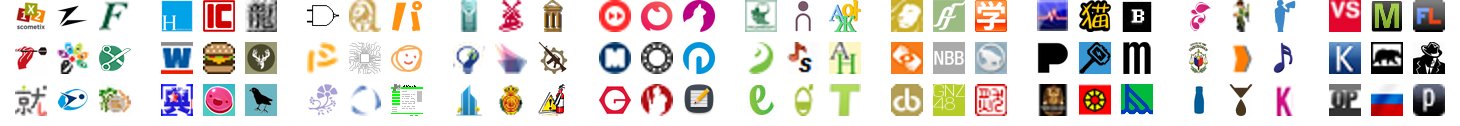}
	\vspace{0.2cm}
	\includegraphics[width=\textwidth]{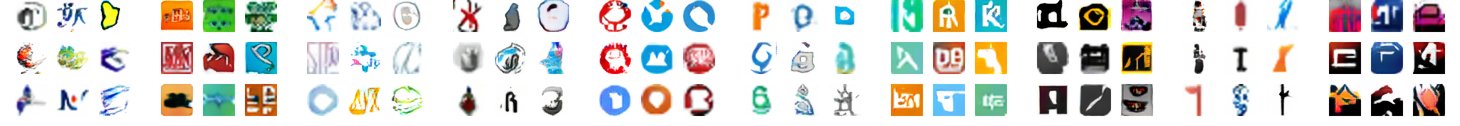}
	\includegraphics[width=\textwidth]{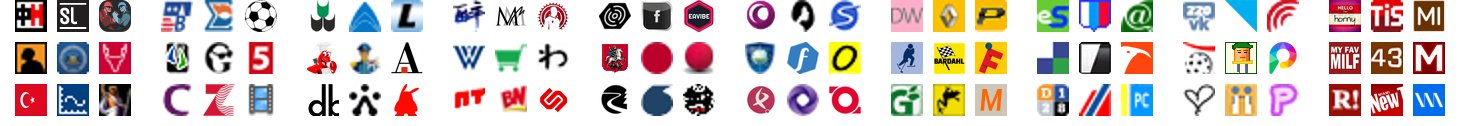}
	\vspace{0.2cm}
	\includegraphics[width=\textwidth]{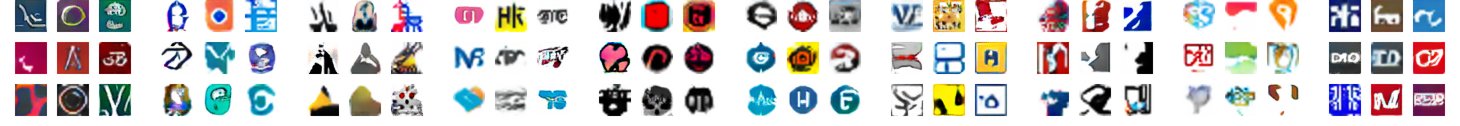}
	\includegraphics[width=\textwidth]{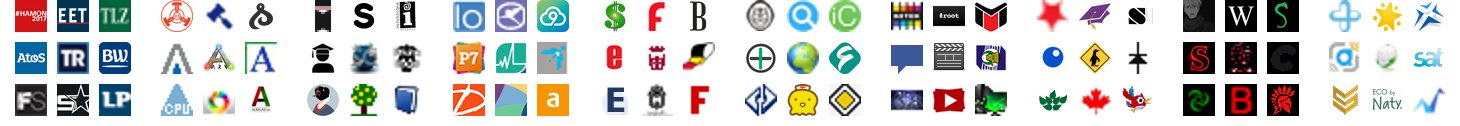}
	\vspace{0.2cm}
	\includegraphics[width=\textwidth]{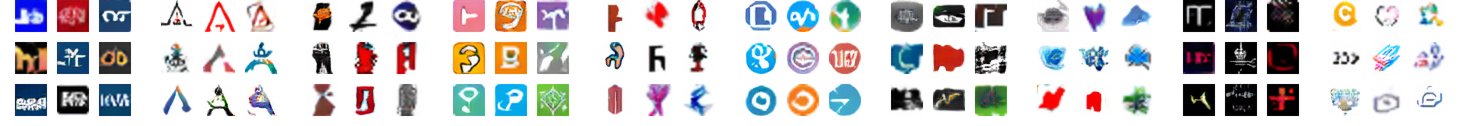}
	\includegraphics[width=\textwidth]{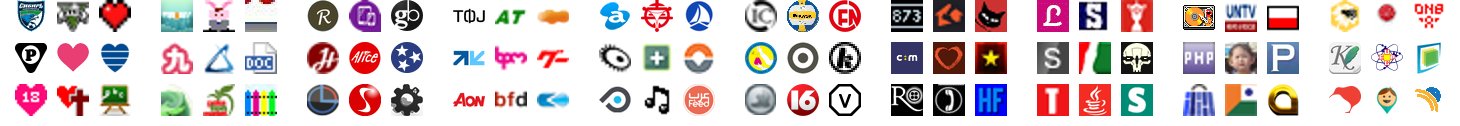}
	\vspace{0.2cm}
	\includegraphics[width=\textwidth]{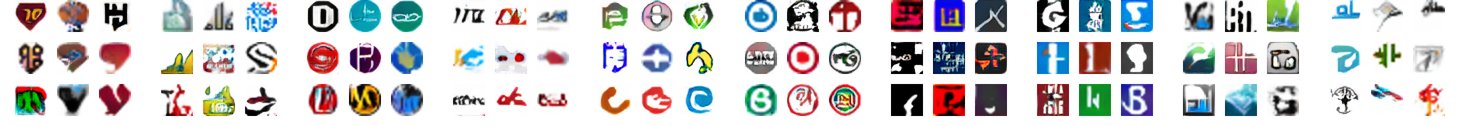}
	\includegraphics[width=\textwidth]{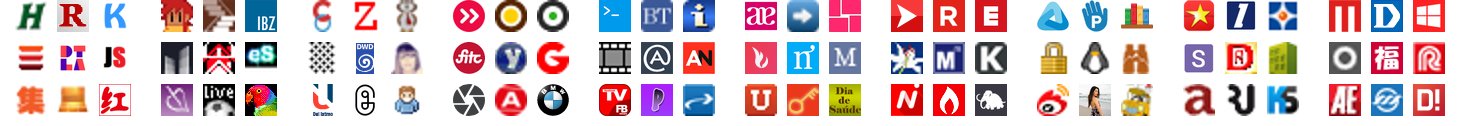}
	\vspace{0.2cm}
	\includegraphics[width=\textwidth]{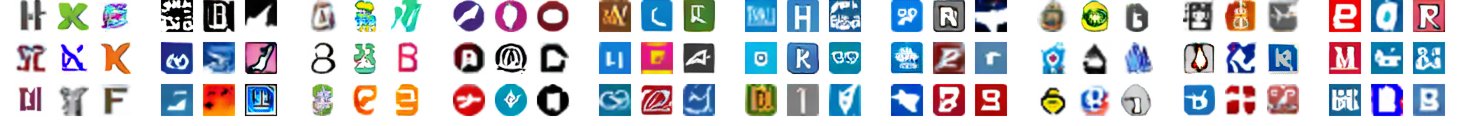}
	\includegraphics[width=\textwidth]{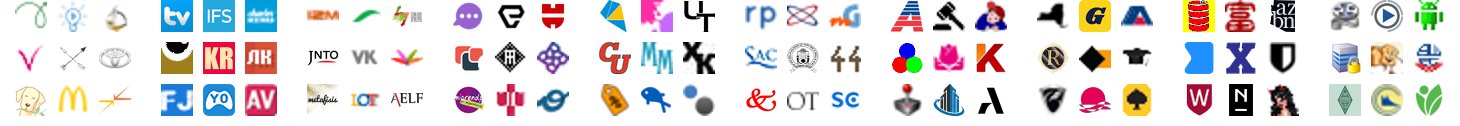}
	\vspace{0.2cm}
	\includegraphics[width=\textwidth]{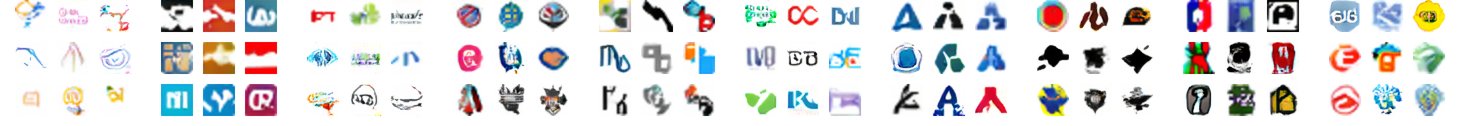}\\
	\vspace{-0.2cm}
	\caption{Clusters 1-70 of LLD-icon clustered in the latent space of an Autoencoder with 100 cluster centers. The top half of each block contains 9 random samples of original images from the cluster, while the bottom half contains 9 random samples from the DCGAN-LC Generator.}
	\label{fig:LLD-icon-ae-cl1}
\end{figure*}

\begin{figure*}[tbp]
	\centering
	\includegraphics[width=\textwidth]{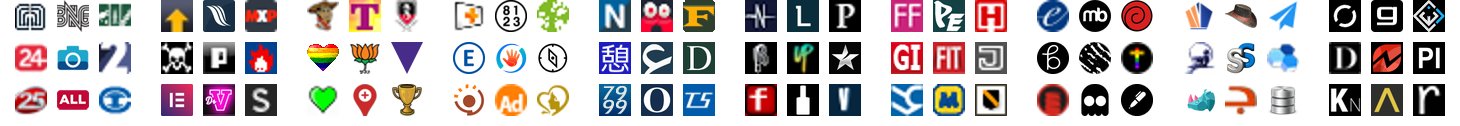}
	\vspace{0.2cm}
	\includegraphics[width=\textwidth]{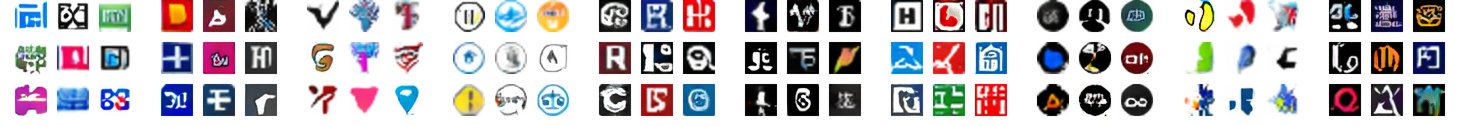}
	\includegraphics[width=\textwidth]{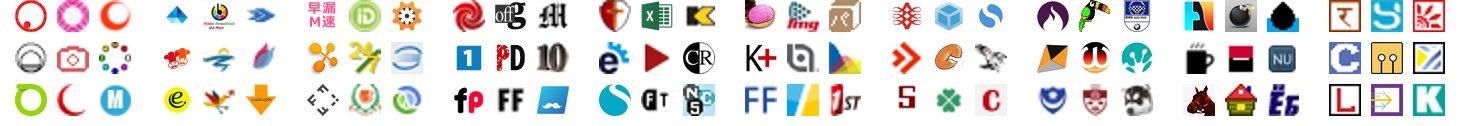}
	\vspace{0.2cm}
	\includegraphics[width=\textwidth]{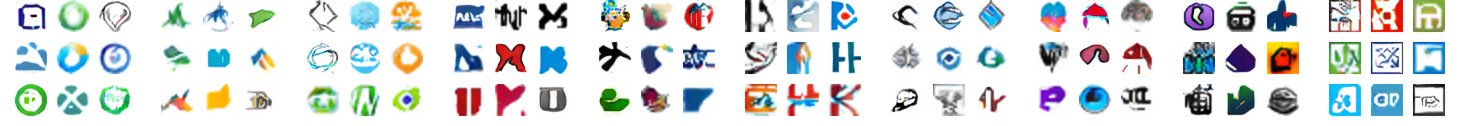}
	\includegraphics[width=\textwidth]{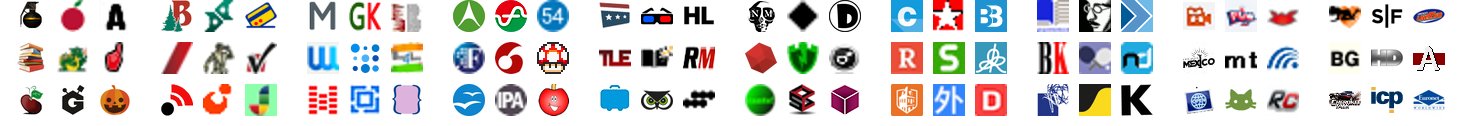}
	\vspace{0.2cm}
	\includegraphics[width=\textwidth]{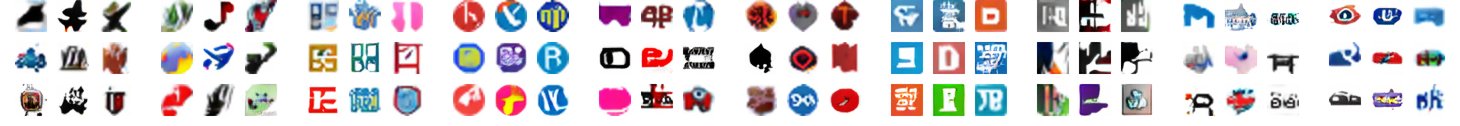}
	\caption{Clusters 71-128 of LLD-icon clustered in the latent space of an Autoencoder with 100 cluster centers. The top half of each block contains 9 random samples of original images from the cluster, while the bottom half contains 9 random samples from the DCGAN-LC Generator.}
	\label{fig:LLD-icon-ae-cl2}
\end{figure*}

\begin{figure*}[tbp]
	\centering
	\vspace{-0.5cm}
	\includegraphics[width=\textwidth]{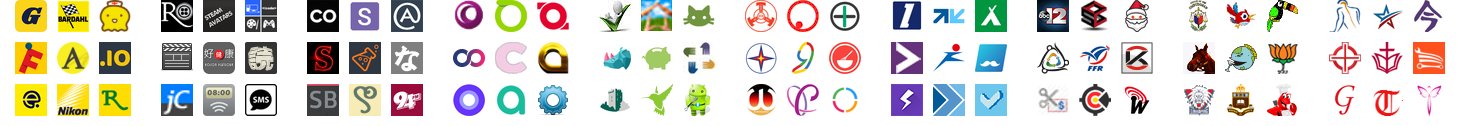}
	\vspace{0.2cm}
	\includegraphics[width=\textwidth]{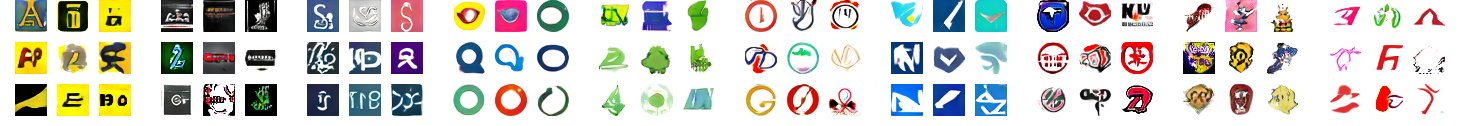}
	\includegraphics[width=\textwidth]{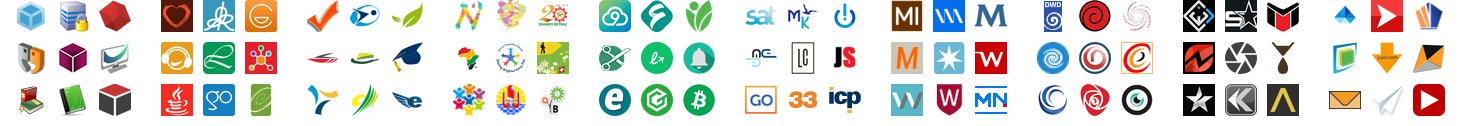}
	\vspace{0.2cm}
	\includegraphics[width=\textwidth]{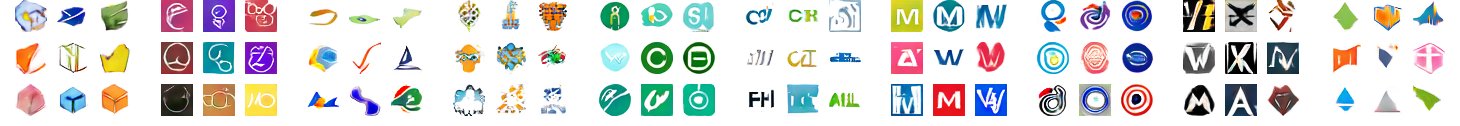}
	\includegraphics[width=\textwidth]{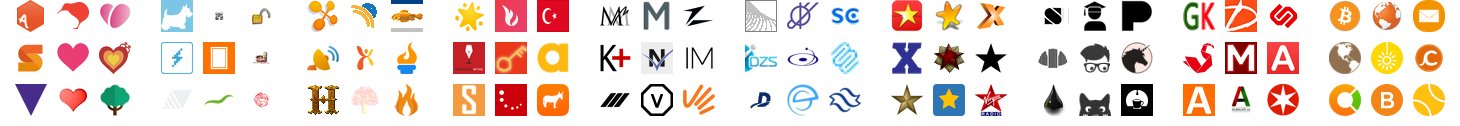}
	\vspace{0.2cm}
	\includegraphics[width=\textwidth]{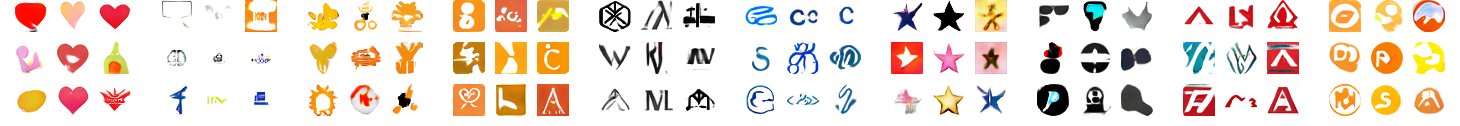}
	\includegraphics[width=\textwidth]{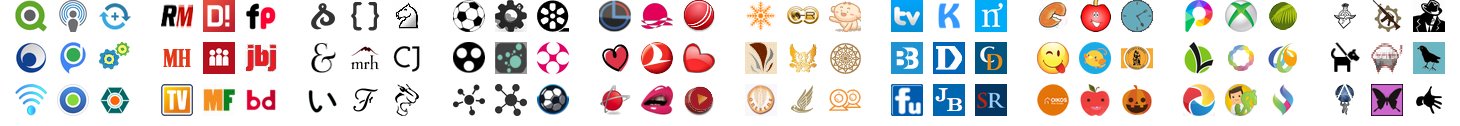}
	\vspace{0.2cm}
	\includegraphics[width=\textwidth]{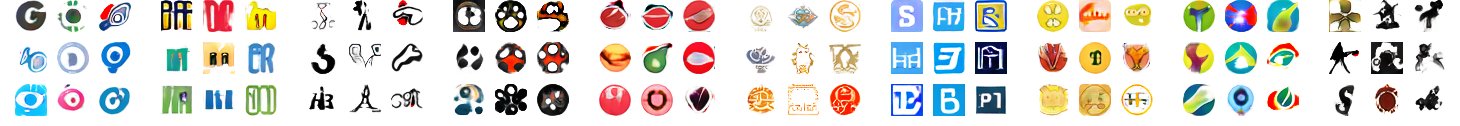}
	\includegraphics[width=\textwidth]{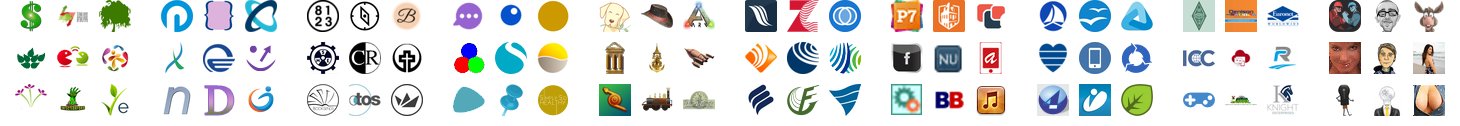}
	\vspace{0.2cm}
	\includegraphics[width=\textwidth]{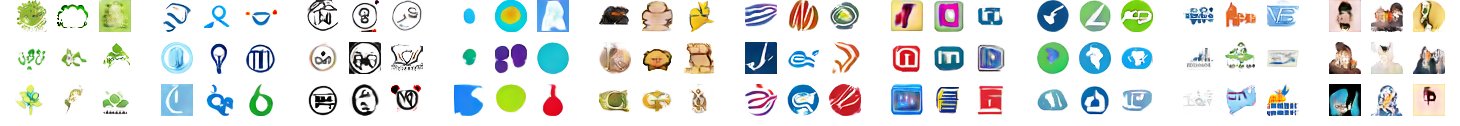}
	\includegraphics[width=\textwidth]{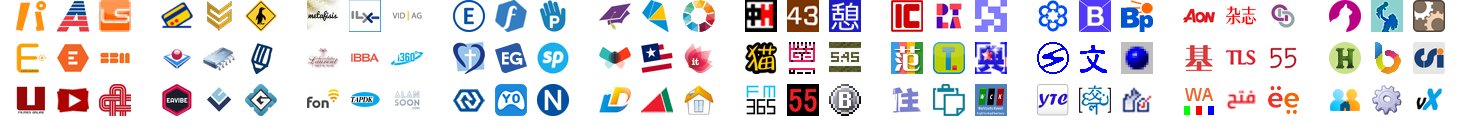}
	\vspace{0.2cm}
	\includegraphics[width=\textwidth]{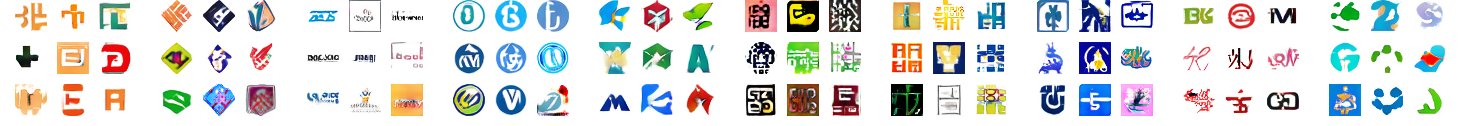}
	\includegraphics[width=\textwidth]{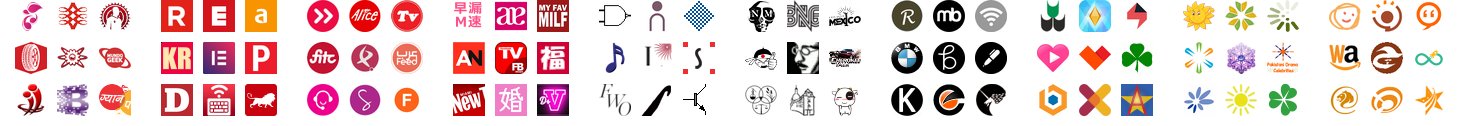}
	\vspace{0.2cm}
	\includegraphics[width=\textwidth]{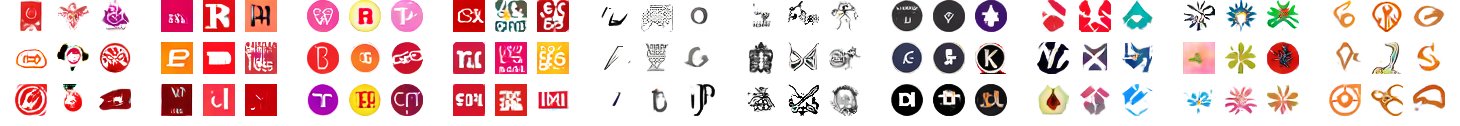}
	\vspace{-0.5cm}
	\caption{Clusters 71-100 of LLD-icon clustered in the latent space of an Autoencoder with 100 cluster centers. The top half of each block contains 9 random samples of original images from the cluster, while the bottom half contains 9 random samples from the DCGAN-LC Generator.}
	\label{fig:LLD-icon-sharp-cl1}
\end{figure*}

\begin{figure*}[tbp]
	\centering
	\includegraphics[width=\textwidth]{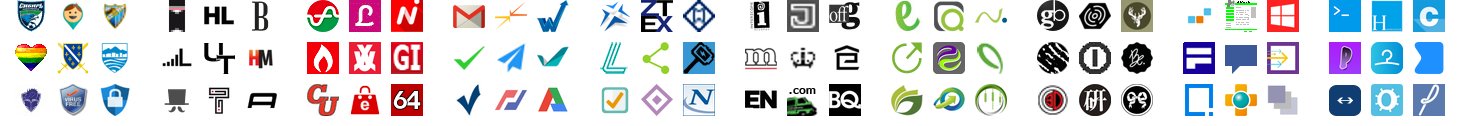}
	\vspace{0.2cm}
	\includegraphics[width=\textwidth]{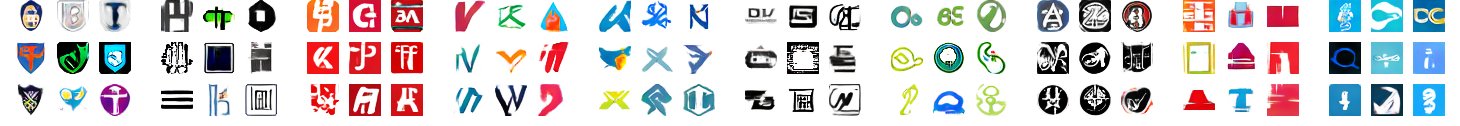}
	\includegraphics[width=\textwidth]{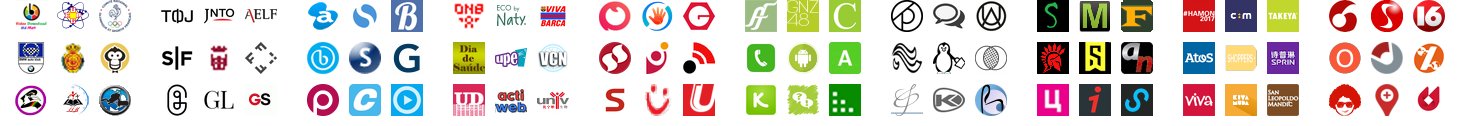}
	\vspace{0.2cm}
	\includegraphics[width=\textwidth]{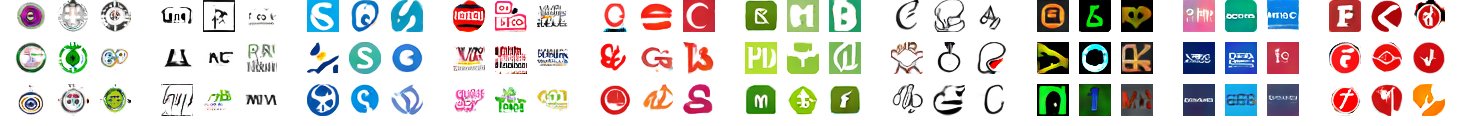}
	\includegraphics[width=\textwidth]{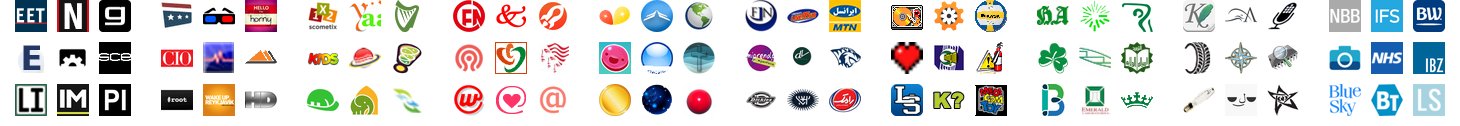}
	\vspace{0.2cm}
	\includegraphics[width=\textwidth]{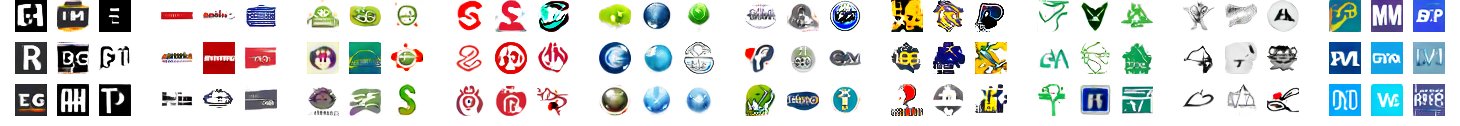}
	\includegraphics[width=\textwidth]{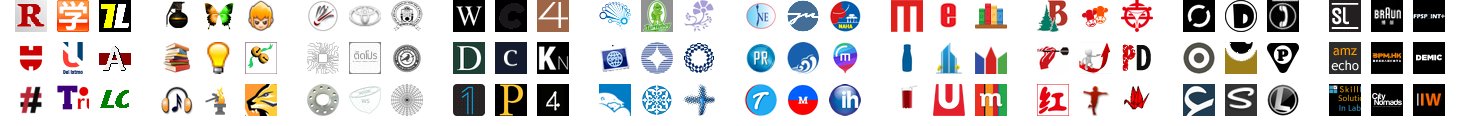}
	\vspace{0.2cm}
	\includegraphics[width=\textwidth]{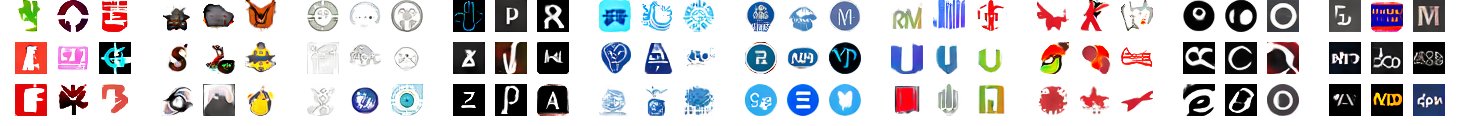}
	\includegraphics[width=\textwidth]{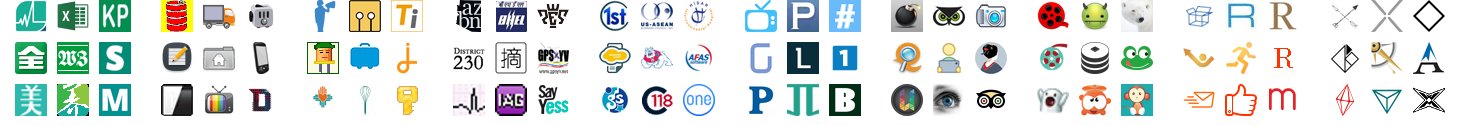}
	\vspace{0.2cm}
	\includegraphics[width=\textwidth]{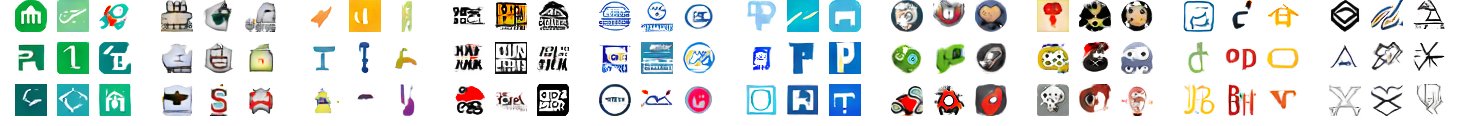}
	\includegraphics[width=\textwidth]{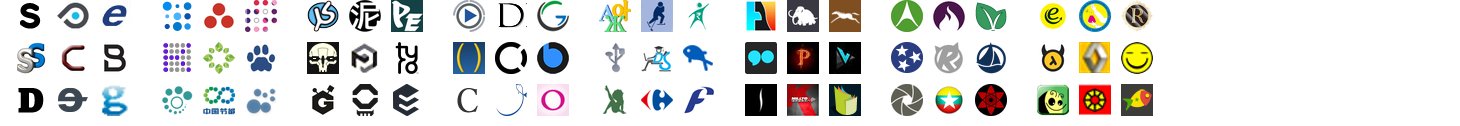}
	\vspace{0.2cm}
	\includegraphics[width=\textwidth]{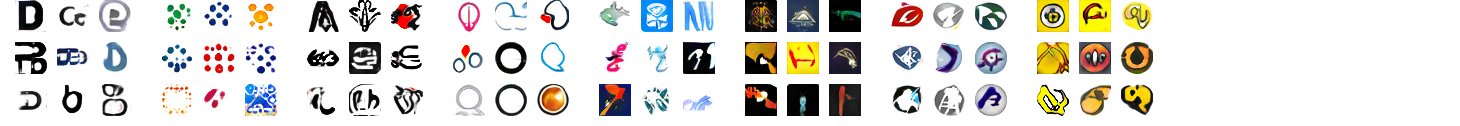}
	\caption{Clusters 71-128 of LLD-icon-sharp clustered with a ResNet Classifier and 128 cluster centers. The top half of each block contains 9 random samples of original images from the cluster, while the bottom half contains 9 random samples from the iWGAN-LC Generator.}
	\label{fig:LLD-icon-sharp-cl2}
\end{figure*}

In a similar way, Figures~\ref{fig:LLD-icon-clean} and \ref{fig:LLD-icon-sharp} present samples from LLD-icon and LLD-icon-sharp, respectively. Here we compare random samples from different trained models, containing both conditional and unconditional variants.
Figure~\ref{fig:LLD-icon-ae-cl1},~\ref{fig:LLD-icon-ae-cl2}, show the clusters found in LLD-icon by clustering in the latent space of an Autoencoder, while Figures~\ref{fig:LLD-icon-sharp-cl1} and~\ref{fig:LLD-icon-sharp-cl2} show clusters in LLD-icon-sharp from the feature-space of a ResNet classifier. A very noticeable difference originates from the fact that the Autoencoder was trained on gray-scale images and is thus relatively color-independent, while there are some very apparent single-color clusters in the RC-version, mostly containing green, blue or orange/red logos.

\begin{figure*}[tbp]
	\centering
	\begin{subfloat}[iWgan unconditional. Inception score: 7.85]{
			\centering
			\includegraphics[width=0.48\textwidth]{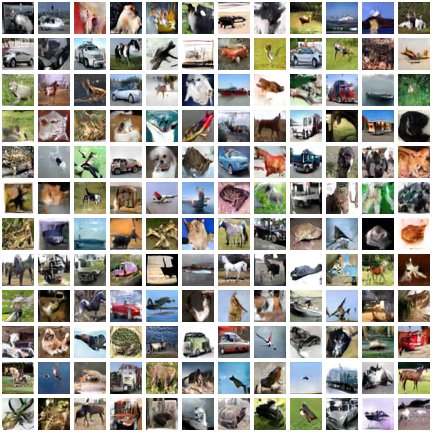}
		}
	\end{subfloat}
	\begin{subfloat}[iWGAN-AC with 32 RC clusters. Inception score: 8.67]{
			\centering
			\includegraphics[width=0.48\textwidth]{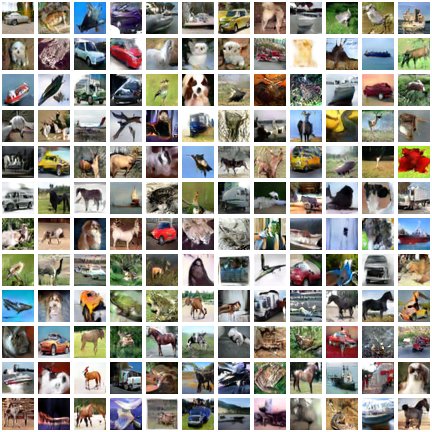}
		}
	\end{subfloat}
	\begin{subfloat}[iWGAN-AC with original labels. Inception score: 8.35]{
			\centering
			\includegraphics[width=0.48\textwidth]{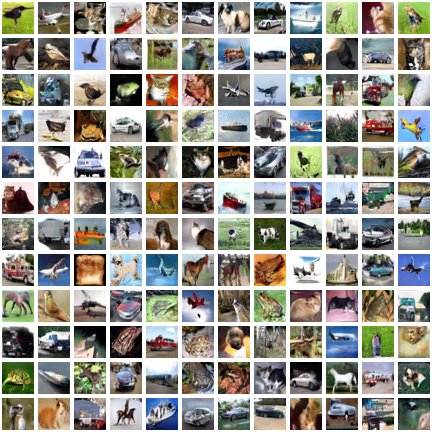}
		}
	\end{subfloat}
	\begin{subfloat}[iWGAN-LC with 32 RC clusters. Inception score: 7.83]{
			\centering
			\includegraphics[width=0.48\textwidth]{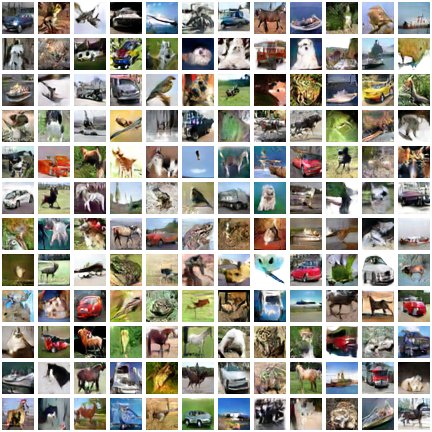}
		}
	\end{subfloat}
	\caption{Random samples from different iWGAN models trained on CIFAR-10 data.}
	\label{fig:CIFAR-10-generators}
\end{figure*}

\begin{figure*}[tbp]
	\centering
	\vspace{-0.5cm}
	\begin{subfloat}[Original data labels (10 categories)]{
			\centering
			\includegraphics[width=\textwidth]{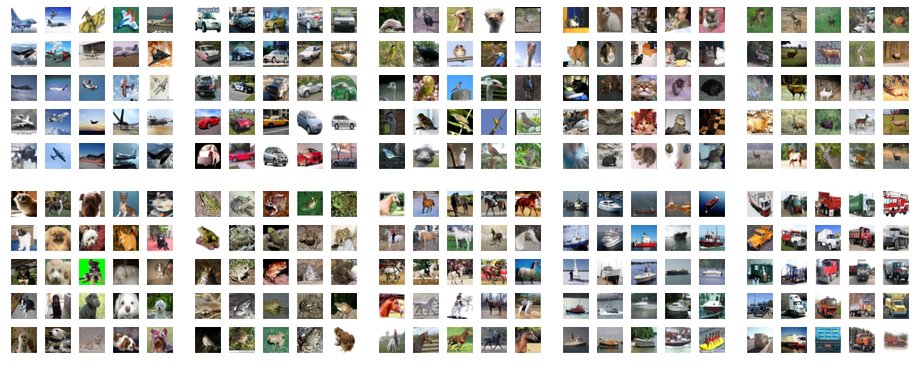}
		}
	\end{subfloat}
	\begin{subfloat}[Clustering in Autoencoder space with 32 cluster centers]{
			\centering
			\includegraphics[width=\textwidth]{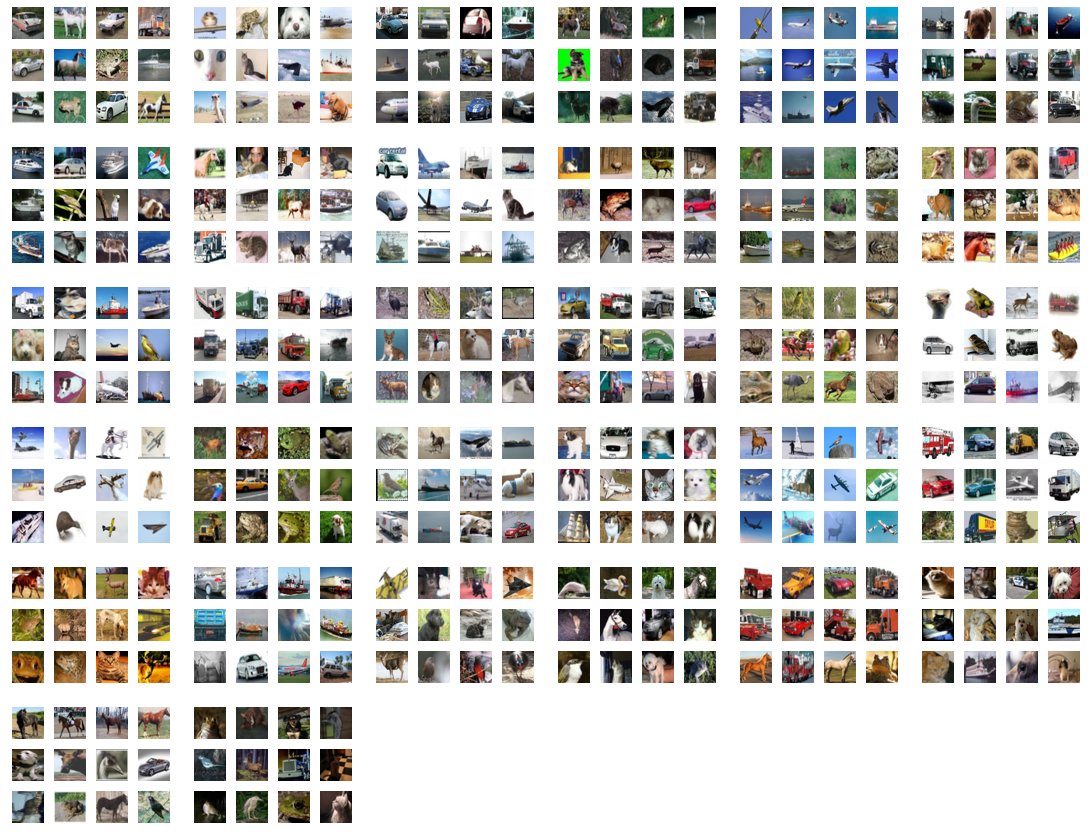}
		}
	\end{subfloat}
	\caption{Original labels and 32 AE clusters. Note the strong variability in visual appearance within the semantic classes, pointing to a possible advantage of using a clustering more in-line with visual semantics. Our experiments with AE clustering produced clearly inferior results on the CIFAR-10 dataset (as compared to our own LLD data).}
	\label{fig:cifar-clusters1}
\end{figure*}

\begin{figure*}[tbp]
	\centering
	\vspace{-0.5cm}
	\begin{subfloat}[Clustering in the CNN feature space of a ResNet classifier with 10 cluster centers]{
			\centering
			\includegraphics[width=\textwidth]{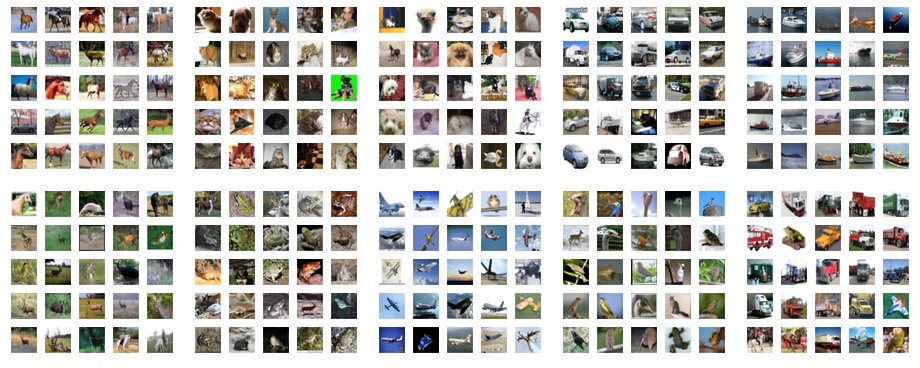}
		}
	\end{subfloat}
	\begin{subfloat}[Clustering in the CNN feature space of a ResNet classifier with 32 cluster centers]{
			\centering
			\includegraphics[width=\textwidth]{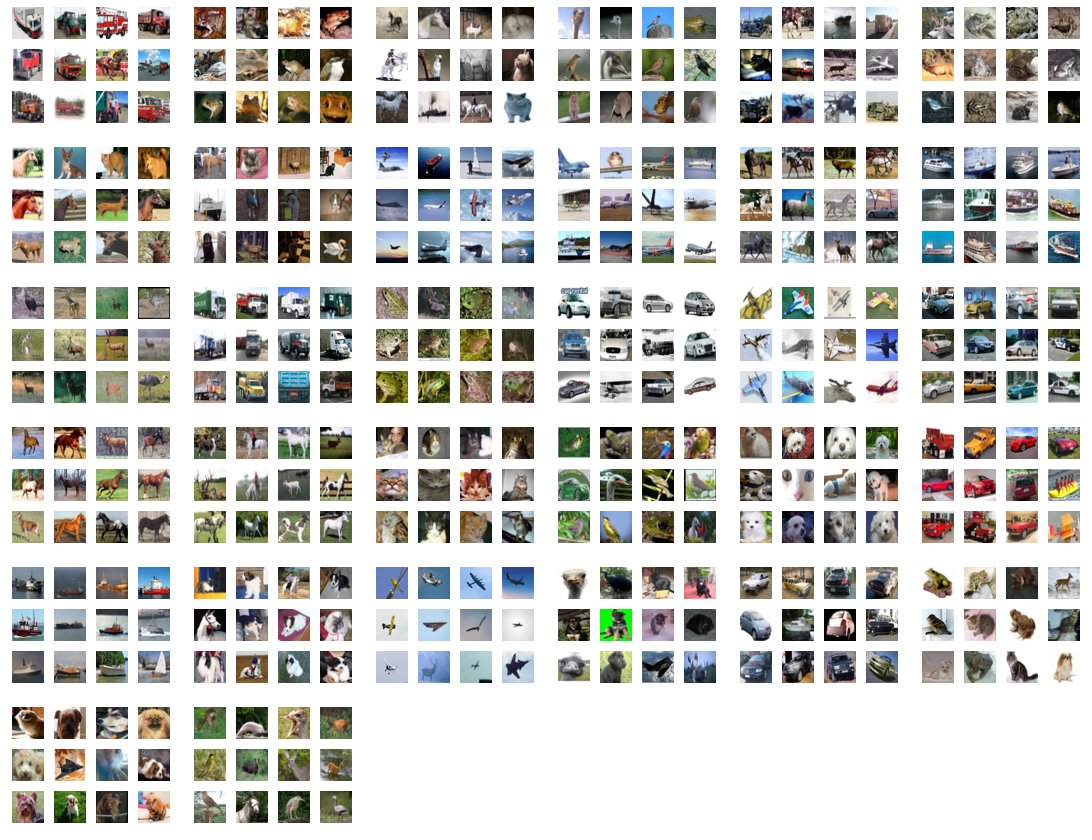}
		}
	\end{subfloat}
	\caption{Resulting clusters using RC clustering with 10 and 32 cluster centers. Compared to the original labels in Figure~\ref{fig:cifar-clusters1}, the 10 clusters shown here are more uniform in visual appearance, however increasing the number of clusters to 32 gives each of them an even more visually consistent appearance.}
	\label{fig:cifar-clusters2}
\end{figure*}

Finally, in Figure~\ref{fig:CIFAR-10-generators}, we present some samples from our benchmarked CIFAR-10 Generators, together with the achieved inception score. Figures~\ref{fig:cifar-clusters1}~and~\ref{fig:cifar-clusters2} compare the clusters found using our RC method with the original data labels, with noticeably more visually uniform classes using our synthetic labeling technique.

\clearpage
\pagebreak
\clearpage
\pagebreak
\thispagestyle{empty}

\clearpage
\pagebreak

\clearpage
\pagebreak
\thispagestyle{empty}
\clearpage
\pagebreak
\thispagestyle{empty}

\clearpage
\pagebreak

\clearpage
\pagebreak

\clearpage
\pagebreak
\section{Architecture Details}
\label{sec:architecture}
In this section we specify the exact architectures and hyper-parameters used to train our models.

\paragraph{iWGAN for 32\x32-pixel output}
We use the residual network architecture designed for CIFAR-10 described in \cite{gulrajani2017improvedWGAN} (Appendix C) for this model. For iWGAN-LC, each stage has an input shape of [$128+k$, ...] where $k$ is the number of classes, i.e. the number of cluster centers used in our clustering approach. All training hyper-parameters remain untouched and we never use normalization in the Discriminator as this resulted in consistently superior Inception scores in our CIFAR-10 experiments. We use the exact same model and training parameters with our LLD-icon dataset.

\paragraph{iWGAN for 64x64-pixel output}
For LLD-logo at 64\x64 pixels again the official TensorFlow implementation by Gulrajani~\etal\cite{gulrajani2017improvedWGAN}\footnote{\url{https://github.com/igul222/improved_wgan_training}}. Again, the input for each stage is extended to have a shape of [$N+k$, ...] where $N$ is the size in the original model and $k$ is the number of classes. The only change we made here is to only use $100,000$ iterations and linearly decay the learning rate over these iterations.

\paragraph{DCGAN}
For DCGAN, we deviate from some hyperparameters used in Taehoon Kim's TensorFlow implementation~\footnote{\url{https://github.com/carpedm20/DCGAN-tensorflow}}, namely:
\begin{itemize}
	\item Higher number of feature maps: (128+$k$, 256+$k$, 512+$k$, 1024+$k$) for the Discriminator layers and (256+$k$, 512+$k$, 1034+$k$, 2048+$k$) for the Generator layers, with $k$ again being the number of classes in the LC version.
	\item For each training iteration of the Discriminator, we train the Generator 3 times
	\item Reduced learning rate of 0.0004 (default: 0.002)
	\item Higher latent space dimensionality of 512 components (default: 100)
	\item Blur input images to Discriminator as detailed in Section~\ref{sec:conditional_GAN} of our paper.
\end{itemize}

\end{document}